\documentclass[runningheads]{llncs}
\usepackage{graphicx}
\usepackage{amsmath,amssymb} 
\usepackage{color,cite}
\usepackage{multirow,bm,subfig}
\usepackage[colorlinks,linkcolor=blue]{hyperref}
\usepackage[width=122mm,left=12mm,paperwidth=146mm,height=193mm,top=12mm,paperheight=217mm]{geometry}

\begin{document}

\newcommand\blfootnote[1]{%
  \begingroup
  \renewcommand\thefootnote{}\footnote{#1}%
  \addtocounter{footnote}{-1}%
  \endgroup
}

\title{PyramidBox: A Context-assisted Single Shot Face Detector.} 

\titlerunning{PyramidBox: A Context-assisted Single Shot Face Detector.}
%

\author{Xu Tang${}^*$ \and Daniel K. Du${}^*$ \and Zeqiang He${}$ \and
Jingtuo Liu$^\dagger$}

%
\authorrunning{Xu Tang, Daniel K. Du, Zeqiang He and Jingtuo Liu}

\institute{Baidu Inc. \\
\email{tangxu02@baidu.com,daniel.kang.du@gmail.com,\{hezeqiang,liujingtuo\}@baidu.com}}
\maketitle              

\blfootnote{$^*$ Equal contribution.}
\blfootnote{$^\dagger$ Corresponding author.}

\begin{abstract}
Face detection has been well studied for many years and one of remaining
challenges is to detect small, blurred and partially occluded faces in uncontrolled
environment. This paper proposes a novel context-assisted single shot face detector,
named \emph{PyramidBox} to handle the hard face detection problem.
Observing the importance of the context, we improve the utilization of contextual information in the following three aspects.
First, we design a novel context anchor to supervise high-level contextual feature learning by a semi-supervised method, which we call it PyramidAnchors.
Second, we propose the Low-level Feature Pyramid Network to combine adequate high-level context semantic feature and Low-level facial feature together, which also allows the PyramidBox to predict faces of all scales in a single shot.
Third, we introduce a context-sensitive structure to increase the capacity of prediction network to improve the final accuracy of output.
In addition, we use the method of Data-anchor-sampling to augment the training samples across different scales, which increases the diversity of training data for smaller faces.
By exploiting the value of context, PyramidBox achieves superior performance among the state-of-the-art over the two common face detection benchmarks, FDDB and WIDER FACE. Our code is available in PaddlePaddle: \href{https://github.com/PaddlePaddle/models/tree/develop/fluid/face_detection}{\url{https://github.com/PaddlePaddle/models/tree/develop/fluid/face_detection}}.
\end{abstract}

\section{Introduction}
\label{sec:intro}
Face detection is a fundamental and essential task in various face applications.
The breakthrough work by Viola-Jones~\cite{Viola2004} utilizes AdaBoost algorithm
with Haar-Like features to train a cascade of face vs. non-face classifiers.
Since that, numerous of subsequent
works~\cite{Brubaker2008,Pham2007,Liao2016,Lowe2004,Yang2014,Zhu2006}
are proposed for improving the cascade detectors.
Then, \cite{Mathias2014,Yan2014,Zhu2012} introduce deformable
part models (DPM) into face detection tasks by modeling the relationship of
deformable facial parts. These methods are mainly based on designed features which are less representable and trained by separated steps.

With the great breakthrough of convolutional neural networks(CNN),
a lot of progress for face detection
has been made in recent years due to utilizing modern CNN-based object detectors,
including R-CNN~\cite{Girshick2014,Girshick2016,Girshick2015,Ren2015},
SSD~\cite{Liu2016}, YOLO~\cite{Redmon2016}, FocalLoss~\cite{Lin2017b} and their extensions~\cite{zhang2017single}.
Benefiting from the powerful deep learning approach and end-to-end optimization,
the CNN-based face detectors have achieved much better performance and provided a new baseline for later methods.
Recent anchor-based detection frameworks aim at detecting hard faces in uncontrolled environment such as WIDER FACE~\cite{Barbu14}.
SSH~\cite{Najibi2017} and S$^3$FD~\cite{Zhang2017} develop scale-invariant networks to
detect faces with different scales from different layers in a single network.
Face R-FCN~\cite{Wang2017} re-weights embedding responses on score maps and eliminates
the effect of non-uniformed contribution in each facial part using a position-sensitive
average pooling. FAN~\cite{Wang2017b} proposes an anchor-level attention by highlighting the features from
the face region to detect the occluded faces.

\begin{figure}[h]
\centering
\includegraphics[height=1.8cm]{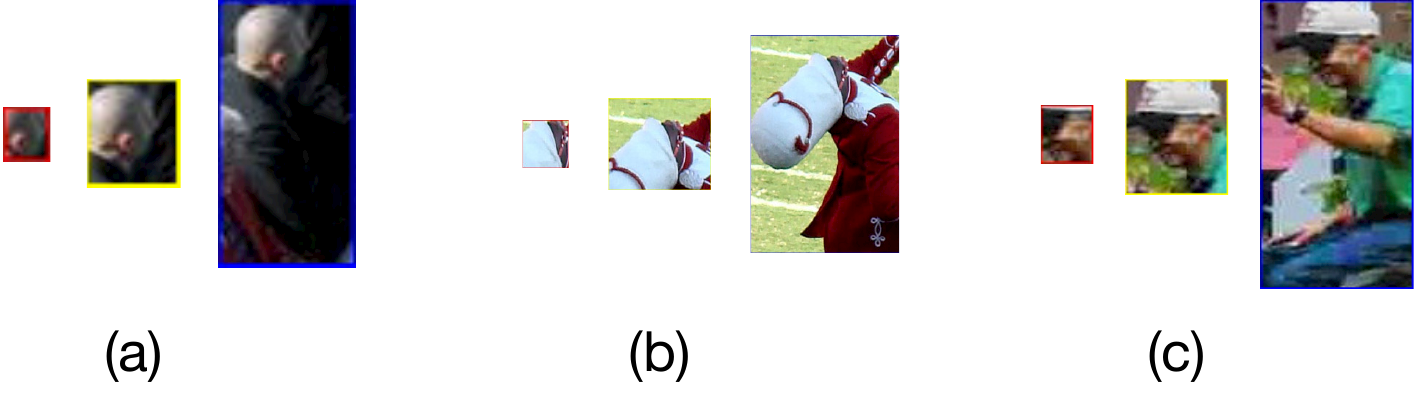}
\caption{Hard faces are difficult to be located and classified due to the lack of visual consistency, while the larger regions which give hints to the position of face are easier to be located and classified, such as head and body.}
\label{fig:context_of_face}
\end{figure}

Though these works give an effective way to design anchors and related networks to detect faces with different scales,
how to use the contextual information in face detection has not been paid enough attention, which should play a significant role in detection of hard faces.
Actually, as shown in Fig.~\ref{fig:context_of_face}, it is clear that faces never occur isolated in the real world, usually with shoulders or bodies, providing a rich
source of contextual associations to be exploited especially when the facial texture is not distinguishable for the sake of low-resolution, blur and occlusion.
We address this issue by introducing a novel framework of context assisted network to make full use of contextual signals as the following steps.

Firstly, the network should be able to learn features for not only faces, but also contextual parts such as heads and bodies. To achieve this goal, extra labels
are needed and the anchors matched to these parts should be designed. In this work, we use a semi-supervised solution to generate approximate labels for contextual
parts related to faces and a series of anchors called PyramidAnchors are invented to be easily added to general anchor-based architectures.

Secondly, high-level contextual features should be adequately combined with the low-level ones. The appearances of hard and easy faces can be quite different, which implies that
not all high-level semantic features are really helpful to smaller targets. We investigate the performance of
Feature Pyramid Networks (FPN) ~\cite{Lin2017} and modify it into a \emph{Low-level Feature Pyramid
Network (LFPN)} to join mutually helpful features together.

Thirdly, the predict branch network should make full use of the joint feature. We introduce the \emph{Context-sensitive prediction module (CPM)} to incorporate
context information around the target face with a wider and deeper network. Meanwhile, we propose a max-in-out layer for the prediction module to further improve the
capability of classification network.

In addition, we propose a training strategy named as \emph{Data-anchor-sampling} to make an adjustment on
the distribution of the training dataset. In order to learn more representable features, the diversity of hard-set samples is important and can be gained by data augmentation across samples.

For clarity, the main contributions of this work can be summarized as five-fold:
\begin{enumerate}
\item We propose an anchor-based context assisted method, called PyramidAnchors,
to introduce supervised information on learning contextual features for small, blurred and partially occluded faces.
\item We design the Low-level Feature Pyramid Networks (LFPN) to merge contextual features and facial features better. Meanwhile, the proposed method can handle faces with different scales well in a single shot.
\item We introduce a context-sensitive prediction module, consisting of a mixed network structure and max-in-out layer to learn accurate location and classification from the merged features.
\item We propose the scale aware Data-anchor-sampling strategy to change the distribution of training samples to put emphasis on smaller faces.
\item We achieve superior performance over state-of-the-art on the common face detection benchmarks
FDDB and WIDER FACE.
\end{enumerate}

The rest of the paper is organized as follows. Section~\ref{sec:relate} provides an
overview of the related works. Section~\ref{sec:pyramidbox} introduces the proposed method.
Section~\ref{sec:exper} presents the experiments and Section~\ref{sec:conclu} concludes the paper.

\section{Related Work}
\label{sec:relate}

\textbf{Anchor-based Face Detectors.}
Anchor was first proposed by Faster R-CNN \cite{Ren2015}, and then it was widely used in both two-stage and
one single shot object detectors.
Then anchor-based object detectors~\cite{Liu2016,Redmon2016} have achieved remarkable progress in recent years.
Similar to FPN~\cite{Lin2017}, Lin~\cite{Lin2017b} uses translation-invariant anchor boxes,
and Zhang~\cite{Zhang2017} designs scales of anchors to ensure that the detector can handle various scales of faces well.
FaceBoxes~\cite{Zhang2017b} introduces anchor densification to ensure different types of anchors
have the same density on the image. S$^3$FD~\cite{Zhang2017} proposed anchor matching strategy to improve the
recall rate of tiny faces.

\textbf{Scale-invariant Face Detectors.}
To improve the performance of face detector to handle faces of different scales,
many state-of-the-art works\cite{Yang2017,Zhang2017,Wang2017b,Najibi2017}
construct different structures in the same framework to detect faces with variant size, where
the high-level features are designed to detect large faces while low-level features for small faces.
In order to integrate high-level semantic feature into low-level layers with higher resolution,
FPN~\cite{Lin2017} proposed a top-down architecture to use high-level semantic feature maps
at all scales. Recently, FPN-style framework achieves great performance on both objection detection~\cite{Lin2017b}
and face detection~\cite{Wang2017b}.

\textbf{Context-associated Face Detectors.} Recently, some works
show the importance of contextual information for face detection,
especially for finding small, blurred and occluded faces.
CMS-RCNN~\cite{Zhu2016} used Faster R-CNN in face detection with body contextual information.
Hu et al.~\cite{Hu2017} trained separate detectors for different scales.
SSH~\cite{Najibi2017} modeled the context information by large filters on each prediction module.
FAN~\cite{Wang2017b} proposed an anchor-level attention, by highlighting the features from
the face region, to detect the occluded faces.

\section{PyramidBox}
\label{sec:pyramidbox}
\begin{figure}[t]
\centering
\includegraphics[height=9.0cm]{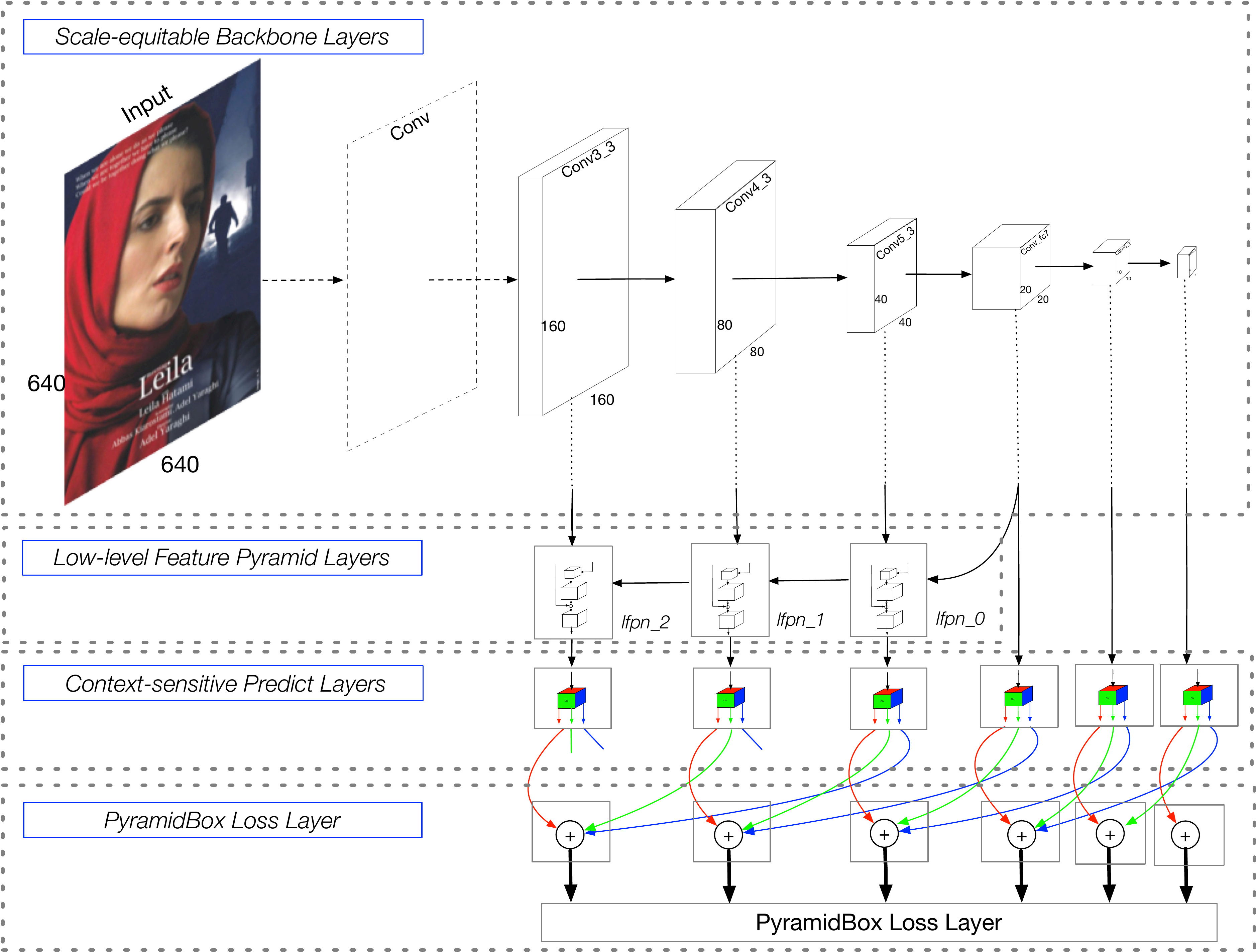}
\caption{Architecture of PyramidBox. It consists of \textbf{Scale-equitable Backbone Layers},
\textbf{Low-level Feature Pyramid Layers (LFPN)},
\textbf{Context-sensitive Predict  Layers}
and \textbf{PyramidBox Loss Layer}.}
\label{fig:architecture}
\end{figure}

\begin{figure}[t]
\centering
\subfloat[]{
 \includegraphics[height=3.5cm]{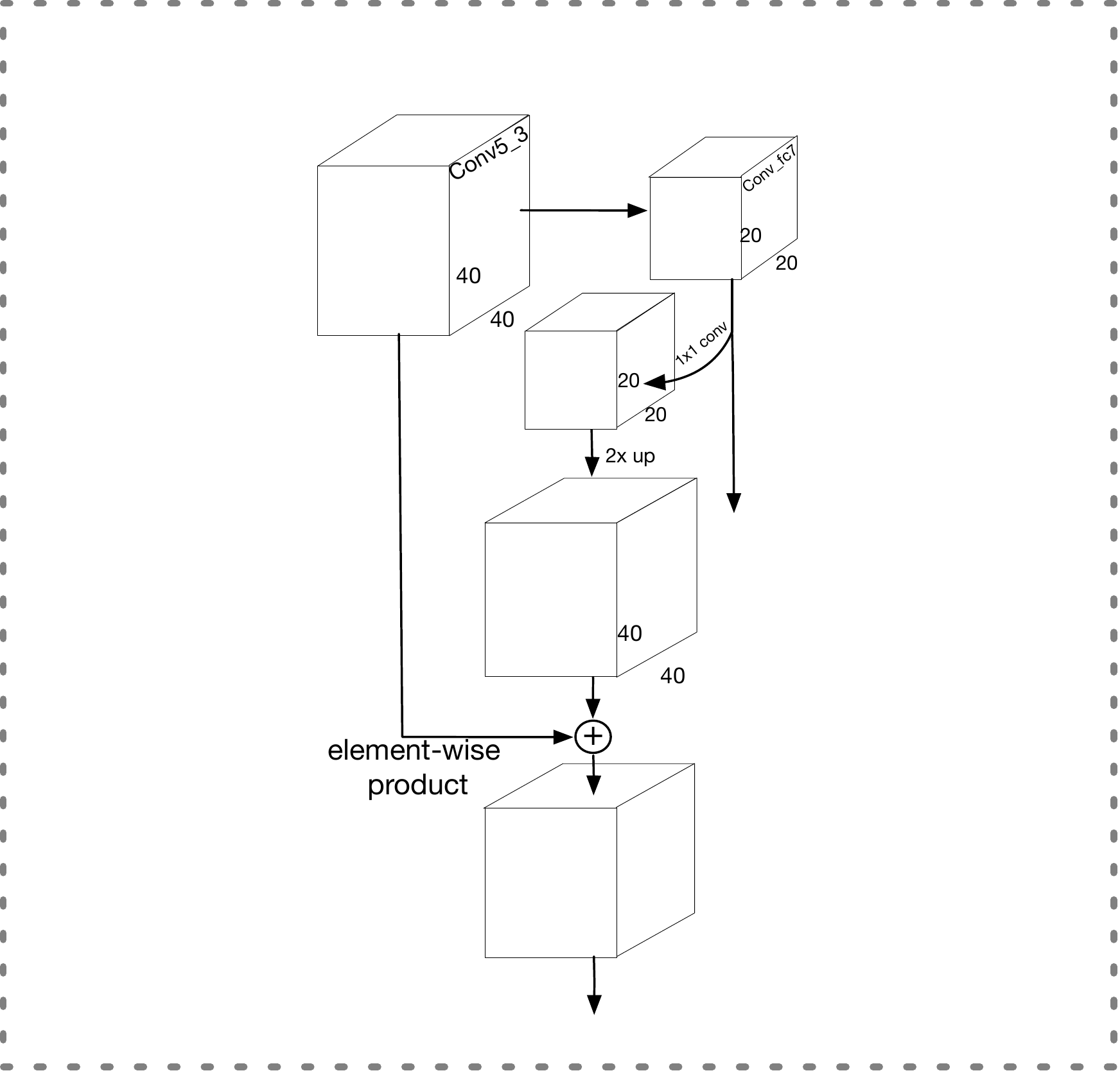}}\hfill
\subfloat[]{
 \includegraphics[height=3.5cm]{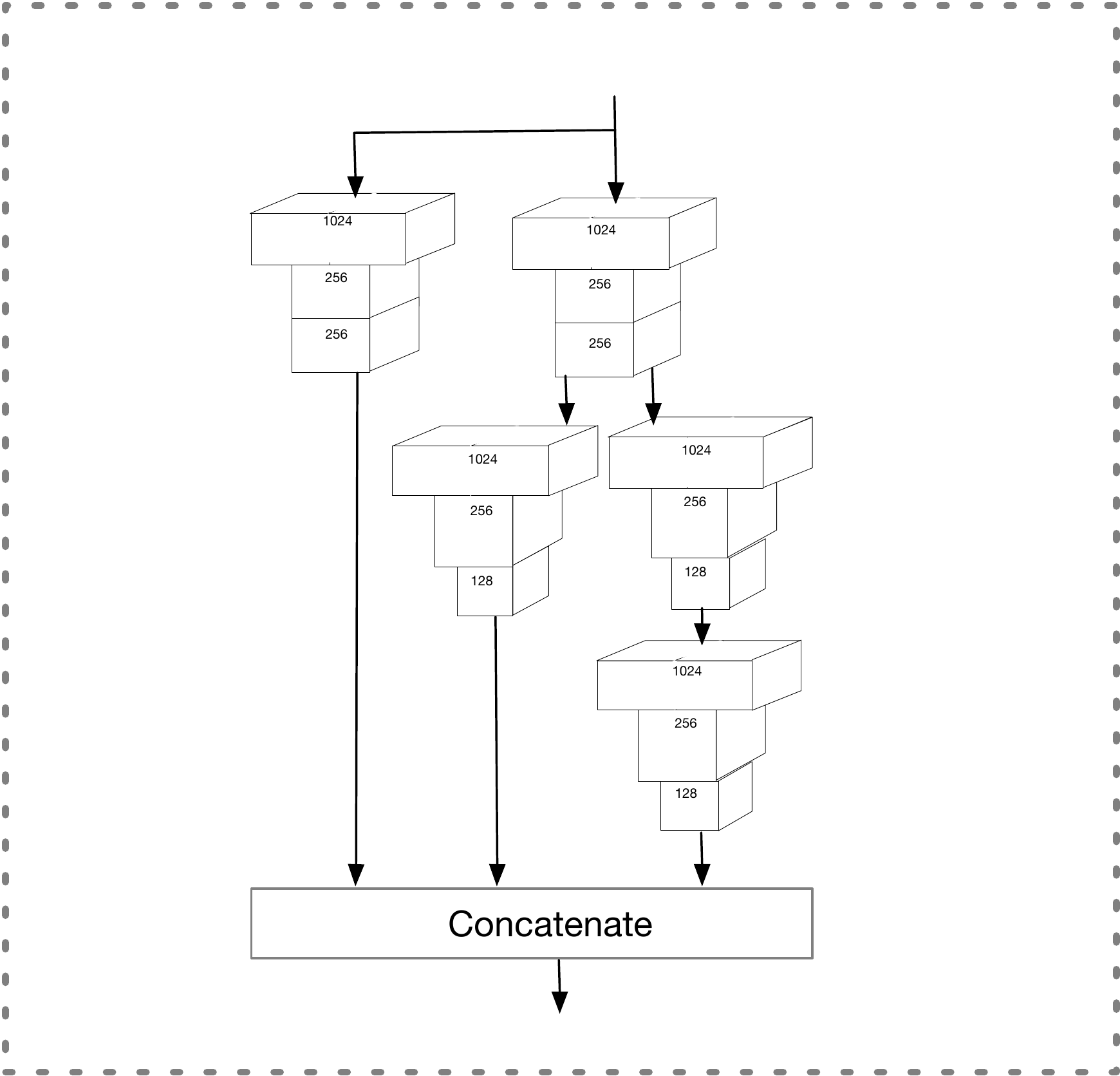}}\hfill
\subfloat[]{
 \includegraphics[height=3.5cm]{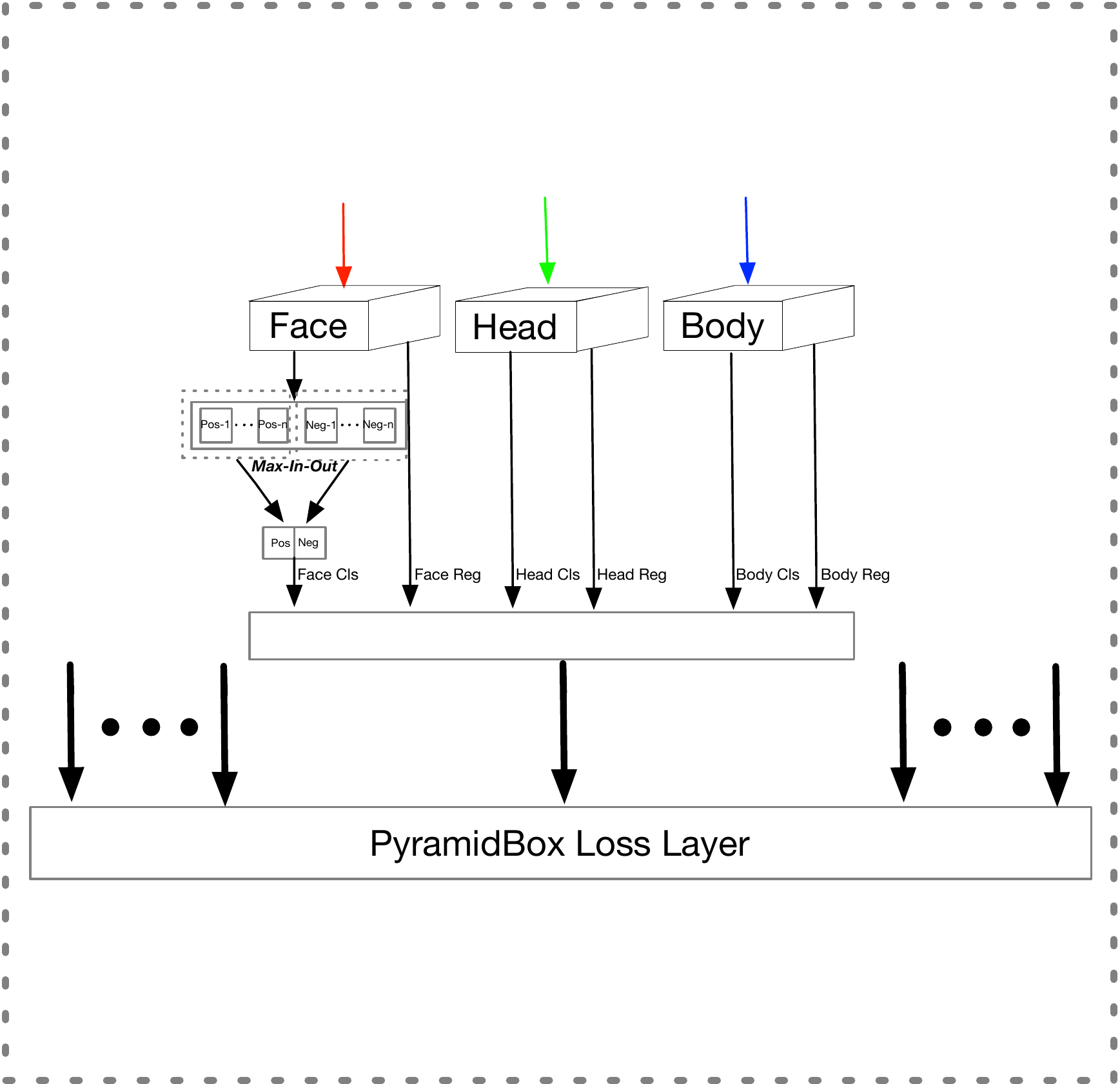}}\hfill
\caption{(a) Feature Pyramid Net. (b) Context-sensitive Prediction Module. (c) PyramidBox Loss.}
\label{fig:architecture_detail}
\end{figure}

This section introduces the context-assisted single shot face detector, \emph{PyramidBox}.
We first briefly introduce
the network architecture in Sec.~\ref{sec:LFPN}.
Then we present a context-sensitive prediction module in Sec.~\ref{sec:cpm},
and propose a novel anchor method, named \emph{PyramidAnchors}, in Sec.~\ref{sec:pyramidanchors}.
Finally, Sec.~\ref{sec:training} presents the associated training methodology
including data-anchor-sampling and max-in-out.

\subsection{Network Architecture}
\label{sec:LFPN}

Anchor-based object detection frameworks with sophisticated design of anchors have been proved effective to handle faces of variable scales when predictions are made at different
levels of feature map ~\cite{Ren2015,Liu2016,Najibi2017,Zhang2017,Wang2017b}.
Meanwhile, FPN structures showed strength on merging high-level features with the lower ones.
The architecture of PyramidBox(Fig.~\ref{fig:architecture}) uses the same extended VGG16
backbone and anchor scale design as S$^3$FD~\cite{Zhang2017}, which can generate feature maps
at different levels and anchors with equal-proportion interval. Low-level FPN is added on this
backbone and a Context-sensitive Predict Module is used as a branch network from each pyramid
detection layer to get the final output. The key is that we design a novel pyramid anchor method
which generates a series of anchors for each face at different levels.
The details of each component in the architecture are as follows:

\textbf{Scale-equitable Backbone Layers.} We use the base convolution layers and extra convolutional layers
in S$^3$FD~\cite{Zhang2017} as our backbone layers, which keep layers of VGG$16$ from \emph{conv\,$1\_1$}
to \emph{pool\,$5$}, then convert \emph{fc\,$6$} and
\emph{fc\,$7$} of VGG$16$ to \emph{conv$\_$fc} layers,
and then add more convolutional layers to make it deeper.

\textbf{Low-level Feature Pyramid Layers.}
To improve the performance of face detector to handle faces of different scales, the low-level feature with high-resolution plays a key role.
Hence, many state-of-the-art works\cite{Yang2017,Zhang2017,Wang2017b,Najibi2017}
construct different structures in the same framework to detect faces with variant size, where
the high-level features are designed to detect large faces while low-level features for small faces.
In order to integrate high-level semantic feature into low-level layers with higher resolution,
FPN~\cite{Lin2017} proposed a top-down architecture to use high-level semantic feature maps
at all scales. Recently, FPN-style framework achieves great performance on both objection detection~\cite{Lin2017b}
and face detection~\cite{Wang2017b}.

As we know, all of these works build FPN start from the top layer, which should be argued that not all high-level features are undoubtedly helpful to small faces.
First, faces that are small, blurred and occluded have different texture feature from the large, clear and complete ones. So it is rude to directly use all high-level features to enhance the performance on small faces.
Second, high-level features are extracted from regions with little face texture and may introduce noise information.
For example, in the backbone layers of our PyramidBox, the receptive field~\cite{Zhang2017} of the top two
layers \emph{conv\,$7\_2$} and \emph{conv\,$6\_2$} are $724$ and $468$, respectively. Notice that the input size
of training image is $640$, which means that the top two layers contain too much noisy context features, so they may not contribute to detecting medium and small faces.

Alternatively, we build the \emph{Low-level Feature Pyramid Network (LFPN)} starting a top-down structure from a middle layer,
whose receptive field should be close to the half of the input size, instead of the top layer.
Also, the structure of each block of LFPN, as same as FPN~\cite{Lin2017}, one can see
Fig.~\ref{fig:architecture_detail}(a) for details.

\textbf{Pyramid Detection Layers.} We select \emph{lfpn\,$\_$2}, \emph{lfpn\,$\_$1}, \emph{lfpn\,$\_$0},
\emph{conv\,$\_$fc\,$7$}, \emph{conv\,$6\_2$} and \emph{conv\,$7\_2$} as detection layers with anchor size of
$16$, $32$, $64$, $128$, $256$ and $512$, respectively.
Here \emph{lfpn\,$\_$2}, \emph{lfpn\,$\_$1} and \emph{lfpn\,$\_$0} are output layer of LFPN based on \emph{conv\,$3\_3$}, \emph{conv\,$4\_3$} and \emph{conv\,$5\_3$}, respectively.
Moreover, similar to other SSD-style methods, we use L$2$ normalization~\cite{Liu2016b} to rescale the norm of
LFPN layers.

\textbf{Predict Layers.} Each detection layer is followed by a \emph{Context-sensitive Predict Module (CPM)},
see Sec~\ref{sec:cpm}.
Notice that the outputs of CPM are used for supervising pyramid anchors, see Sec.~\ref{sec:pyramidanchors}, which
approximately cover face, head and body region in our experiments.
The output size of the $l$-th CPM is $w_l\times h_l \times c_l$,
where $w_l = h_l = 640/2^{2+l}$ is the corresponding
feature size and the channel size $c_l$ equals to $20$ for  $l = 0,1,\ldots,5$. Here the features of
each channels are used for classification and regression of faces, heads and bodies, respectively,
in which the classification of face need $4$ $(= cp_l + cn_l)$ channels, where $cp_l$ and $cn_l$
are max-in-out of foreground and background label respectively, satisfying
\[
cp_l =
 \left\{\begin{array}{ll}
   1, &\mbox{ if $l = 0$,}\\[7pt]
   3, &\mbox{ otherwise.}
 \end{array}\right.
\]
Moreover, the classification of both head and body need two channels, while each of face,
head and body have four channels to localize.

\textbf{PyramidBox loss layers.} For each target face, see in Sec.~\ref{sec:pyramidanchors}, we have a series of
pyramid anchors to supervise the task of classification and regression simultaneously.
We design a \emph{PyramidBox Loss}. see Sec.~\ref{sec:training}, in which we use softmax loss
for classification and smooth L1 loss for regression.

\subsection{Context-sensitive Predict Module}
\label{sec:cpm}

\textbf{Predict Module.} In original anchor-based detectors, such as SSD~\cite{Liu2016} and YOLO~\cite{Redmon2016},
the objective functions are applied to the selected feature maps directly.
As proposed in MS-CNN~\cite{Cai2016}, enlarging the sub-network of each task can improve
accuracy. Recently, SSH~\cite{Najibi2017} increases the receptive field by placing a wider
convolutional prediction module on top of layers with different strides, and DSSD~\cite{Fu2017}
adds residual blocks for each prediction module.
Indeed, both SSH and DSSD make the prediction module deeper and wider separately, so that the prediction module
get the better feature to classify and localize.

Inspired by the Inception-ResNet~\cite{Szegedy2016}, it is quite clear that we can jointly enjoy the gain of wider and deeper network.
We design the \emph{Context-sensitive Predict Module (CPM)}, see Fig.~\ref{fig:architecture_detail}(b),
in which we replace the convolution layers of
context module in SSH by the residual-free prediction module of DSSD.
This would allow our CPM to reap all the benefits of the DSSD module
approach while remaining rich contextual information from SSH context module.

\textbf{Max-in-out.}
The conception of Maxout was first proposed by Goodfellow et al. ~\cite{Goodfellow2013}.
Recently, S$^3$FD~\cite{Zhang2017}
applied max-out background label to reduce the false positive rate of small negatives.
In this work, we use this strategy on both positive and negative samples.
Denote it as max-in-out, see Fig.~\ref{fig:architecture_detail}(c).
We first predict $c_p + c_n$ scores for each prediction module, and then select
$\max{c_p}$ as the positive score. Similarly, we choose the max score of $c_n$ to be the negative score.
In our experiment, we set $c_p = 1$ and $c_n = 3$ for the first prediction module since that small anchors
have more complicated background~\cite{Zhang2017b}, while $c_p = 3$ and $c_n = 1$ for other prediction modules
to recall more faces.

\subsection{PyramidAnchors}
\label{sec:pyramidanchors}
\begin{figure}[t]
\centering
 \includegraphics[height=6cm]{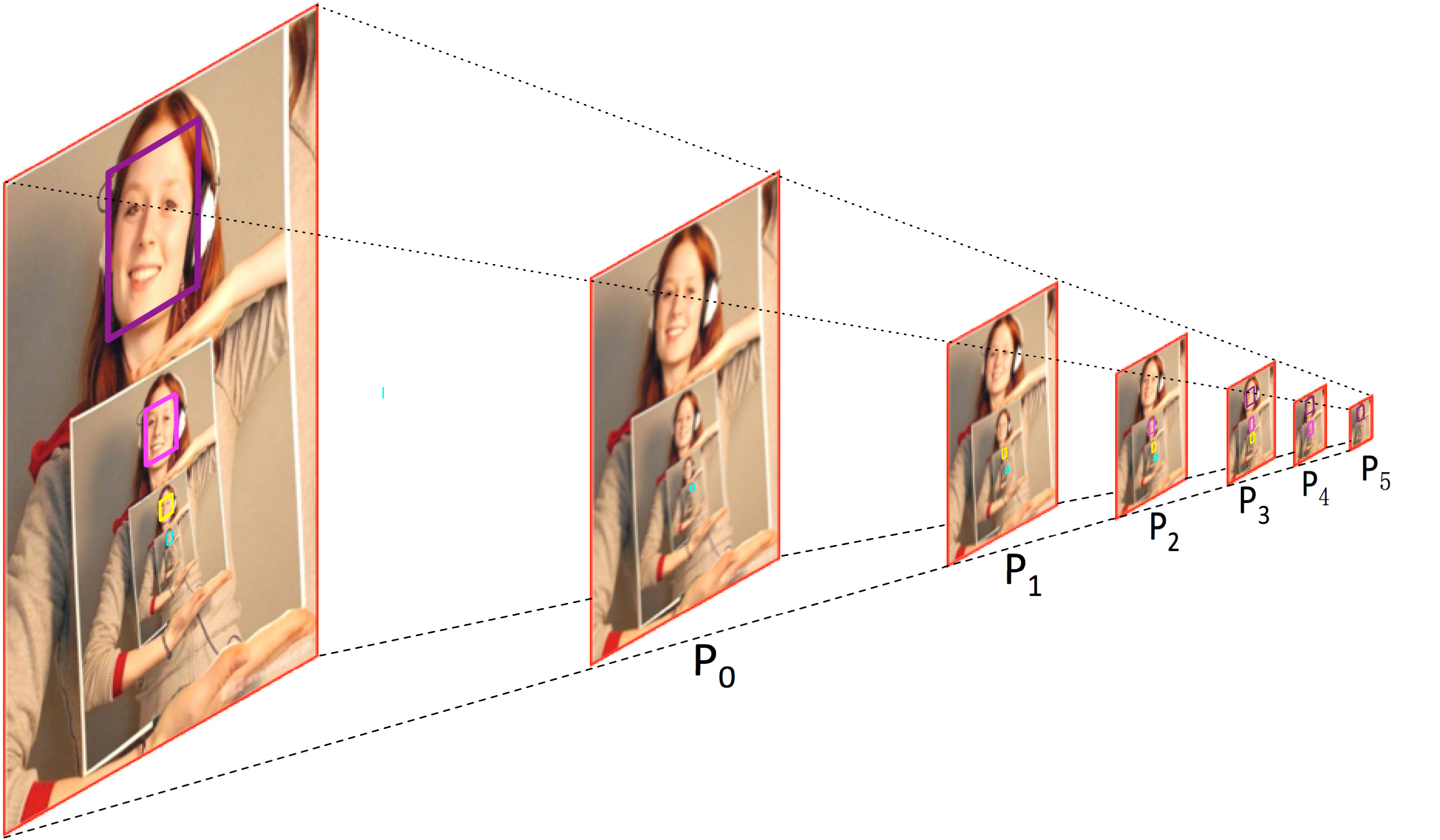}
\caption{Illustration of PyramidAnchors.
For example, the largest purple face with size of $128$
have pyramid-anchors at $P_3$, $P_4$ and $P_5$, where $P_3$ are anchors generated from
\emph{conv$\_fc7$} labeled by the face-self,  $P_4$ are anchors generated from
\emph{conv$6\_2$} labeled by the head (of size about $256$) of the target face,
and  $P_5$ are anchors generated from \emph{conv$7\_2$} labeled by the body (of size about $512$)
of the target face.
Similarly, to detect the smallest cyan face with the size of $16$, one can get a supervised feature from pyramid-anchors
on $P_0$ which labeled by the original face, pyramid-anchors on $P_1$ which labeled by the corresponding
head with size of $32$, and pyramid-anchors on $P_2$ labeled by the corresponding body with
size of $64$. }
\label{fig:pyramidAnchors}
\end{figure}

Recently anchor-based object detectors~\cite{Liu2016,Redmon2016,Lin2017,Lin2017b}
and face detectors~\cite{Zhang2017,Zhang2017b} have achieved remarkable progress.
It has been proved that balanced anchors for each scale are necessary to detect small faces~\cite{Zhang2017}.
But it still ignored the context feature at each scale because the anchors are all designed for face regions.
To address this problem, we propose a novel alternatively anchor method, named \emph{PyramidAnchors}.

For each target face, PyramidAnchors generate a series of anchors corresponding to larger regions
related to a face that contains more contextual information, such as head, shoulder and body.
We choose the layers to set such anchors by matching the region size to the anchor size,
which will supervise higher-level layers to learn more representable features for lower-level scale faces.
Given extra labels of head, shoulder or body, we can accurately match the anchors to ground truth to generate the loss.
As it's unfair to add additional labels, we implement it in a semi-supervised way
under the assumption that regions with the same ratio and offset to different faces own similar contextual feature.
Namely, we can use a set of uniform boxes to approximate the actual regions of head, shoulder and body,
as long as features from these boxes are similar among different faces.
For a target face localized at $region_{target}$ at original image,
considering the $anchor_{i,j}$, which means the $j$-th anchor at
the $i$-th feature layer with stride size $s_i$, we define the label of
$k$-th pyramid-anchor by
\begin{equation}
\label{eq:pyramid_anchor_defi}
label_k(anchor_{i,j}) =
      \left\{ \begin{array}{ll}
              1, & \mbox{if }iou(anchor_{i,j}\cdot s_i/{s_{pa}}^k, region_{target}) > threshold,\\[7pt]
              0, & otherwise,
      \end{array}\right.
\end{equation}
for $k = 0, 1, \ldots, K$, respectively,
where $s_{pa}$ is the stride of pyramid anchors.
$anchor_{i,j}\cdot s_i$ denotes the corresponding region in the original image
of $anchor_{i,j}$, and $anchor_{i,j}\cdot s_i/{s_{pa}}^k$ represents the corresponding
down-sampled region by stride ${s_{pa}}^k$.
The $threshold$ is the same as other anchor-based detectors.
Besides, a PyramidBox Loss will be demonstrated in Sec.~\ref{sec:training}.

In our experiments,
we set the hyper parameter $s_{pa} = 2$ since the stride of adjacent prediction modules is $2$.
Furthermore, let $threshold = 0.35$ and $K=2$.
Then $label_0$, $label_1$ and $label_2$ are labels of face, head and body respectively.
One can see that a face would generate $3$ targets in three continuous prediction modules, which represent
for the face itself, the head and body corresponding to the face.
Fig.~\ref{fig:pyramidAnchors} shows an example.

Benefited from the PyramidBox, our face detector can handle small, blurred and partially occluded faces
better. Notice that the pyramid anchors are generated automatically without
any extra label and this semi-supervised learning help PyramidAnchors extract approximate contextual features.
In prediction process, we only use output of the face branch, so no additional computational
cost is incurred at runtime, compared to standard anchor-based face detectors.

\subsection{Training}
\label{sec:training}
In this section, we introduce the training dataset, data augmentation,
loss function and other implementation details.

\textbf{Train dataset.} We trained PyramidBox on $12,880$ images of the WIDER FACE training set
with color distort,
random crop and horizontal flip.

\begin{figure}[t]
\centering
\subfloat[Pose.]{
\includegraphics[height=2cm]{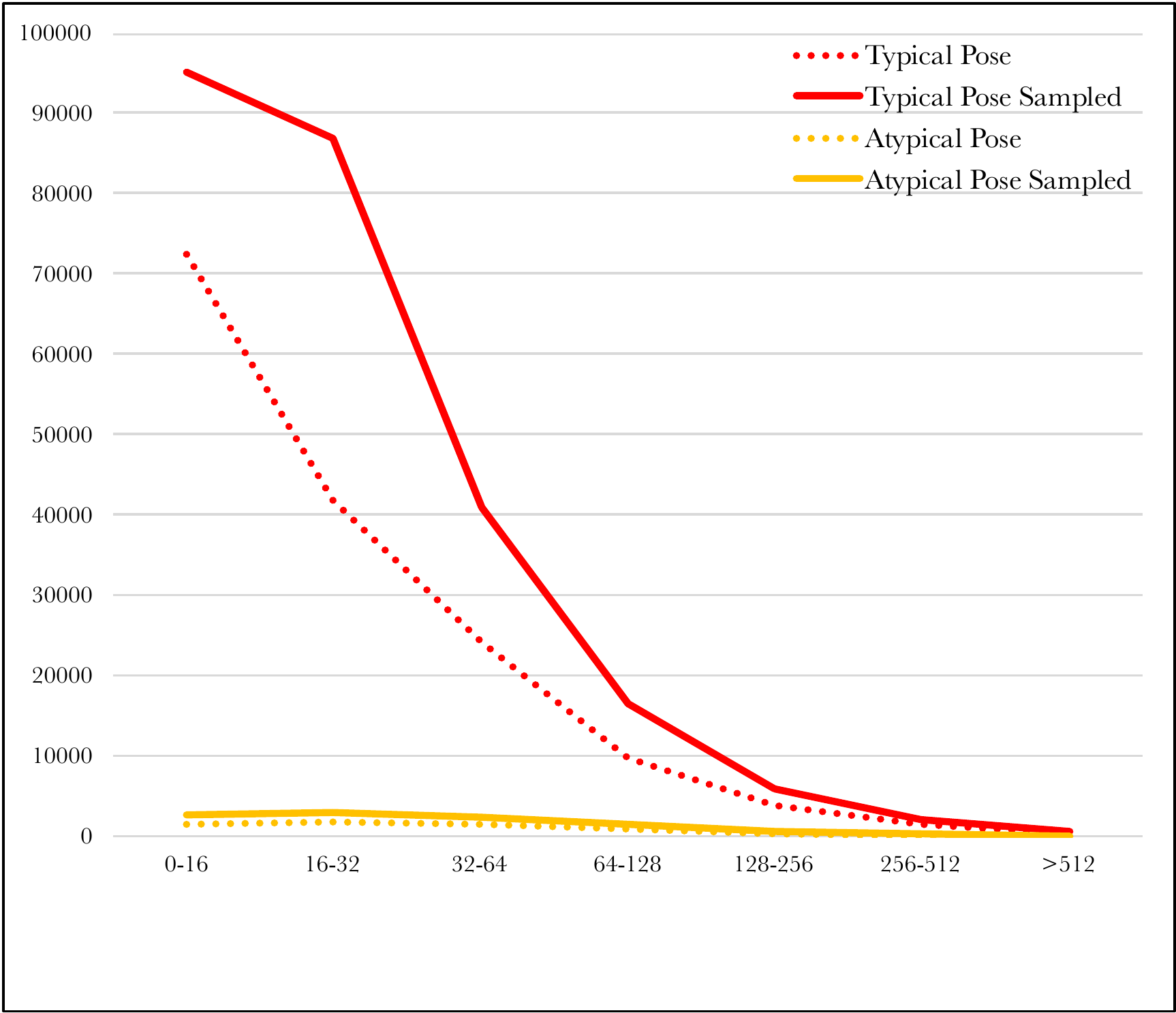}}\hfill
\subfloat[Occlusion.]{
\includegraphics[height=2cm]{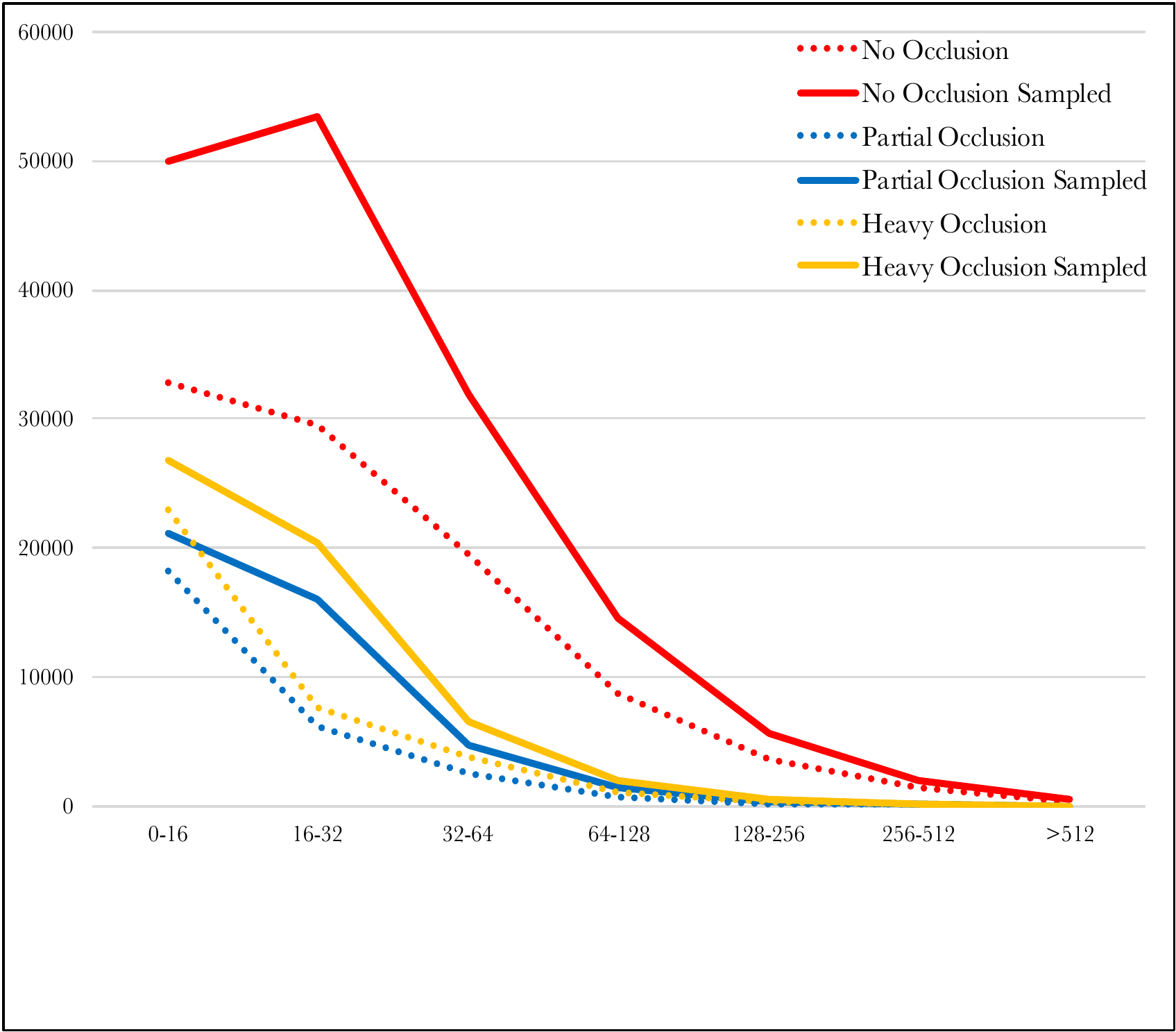}}\hfill
\subfloat[Blur.]{
\includegraphics[height=2cm]{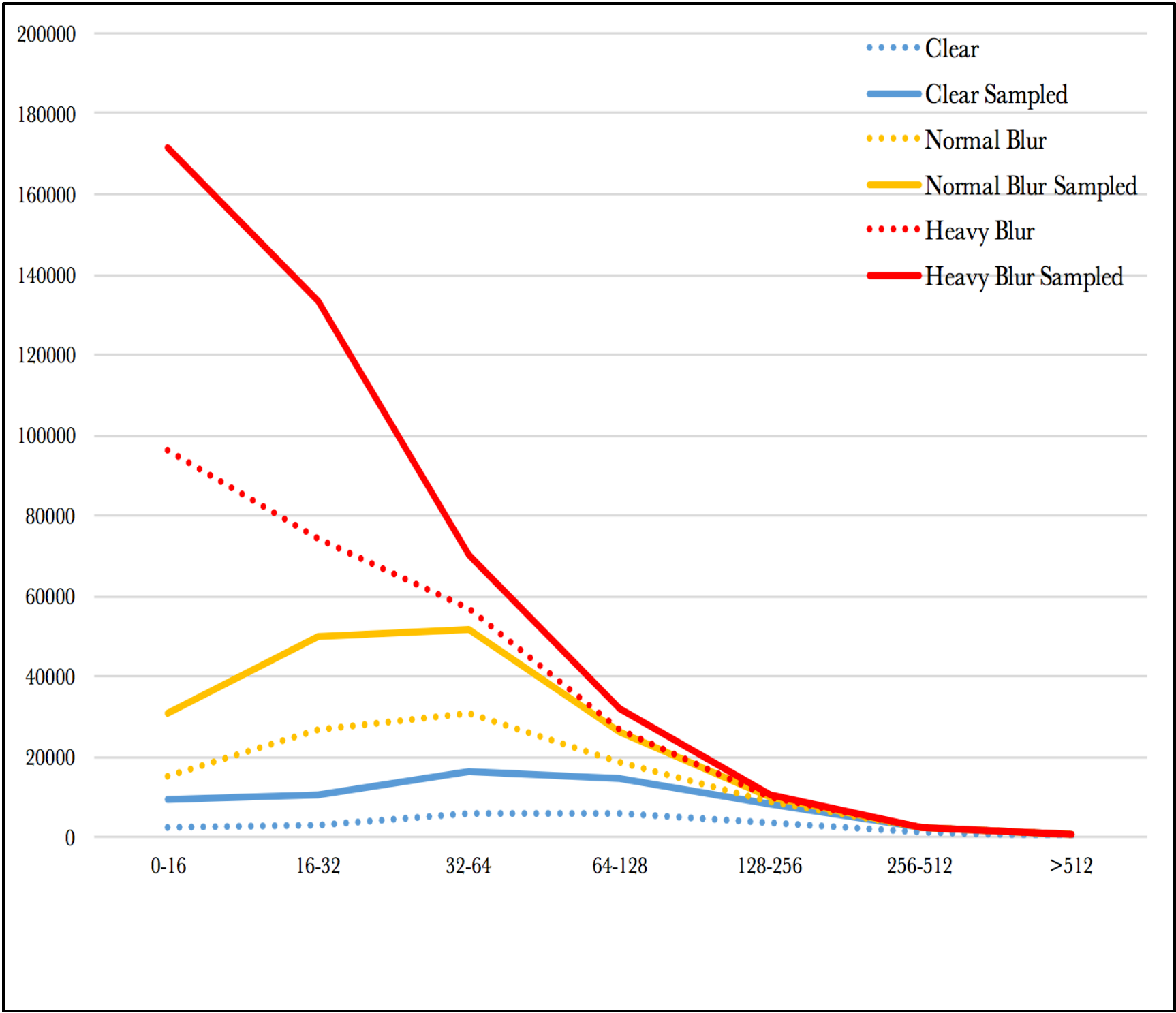}}\hfill
\subfloat[Illumination.]{
\includegraphics[height=2cm]{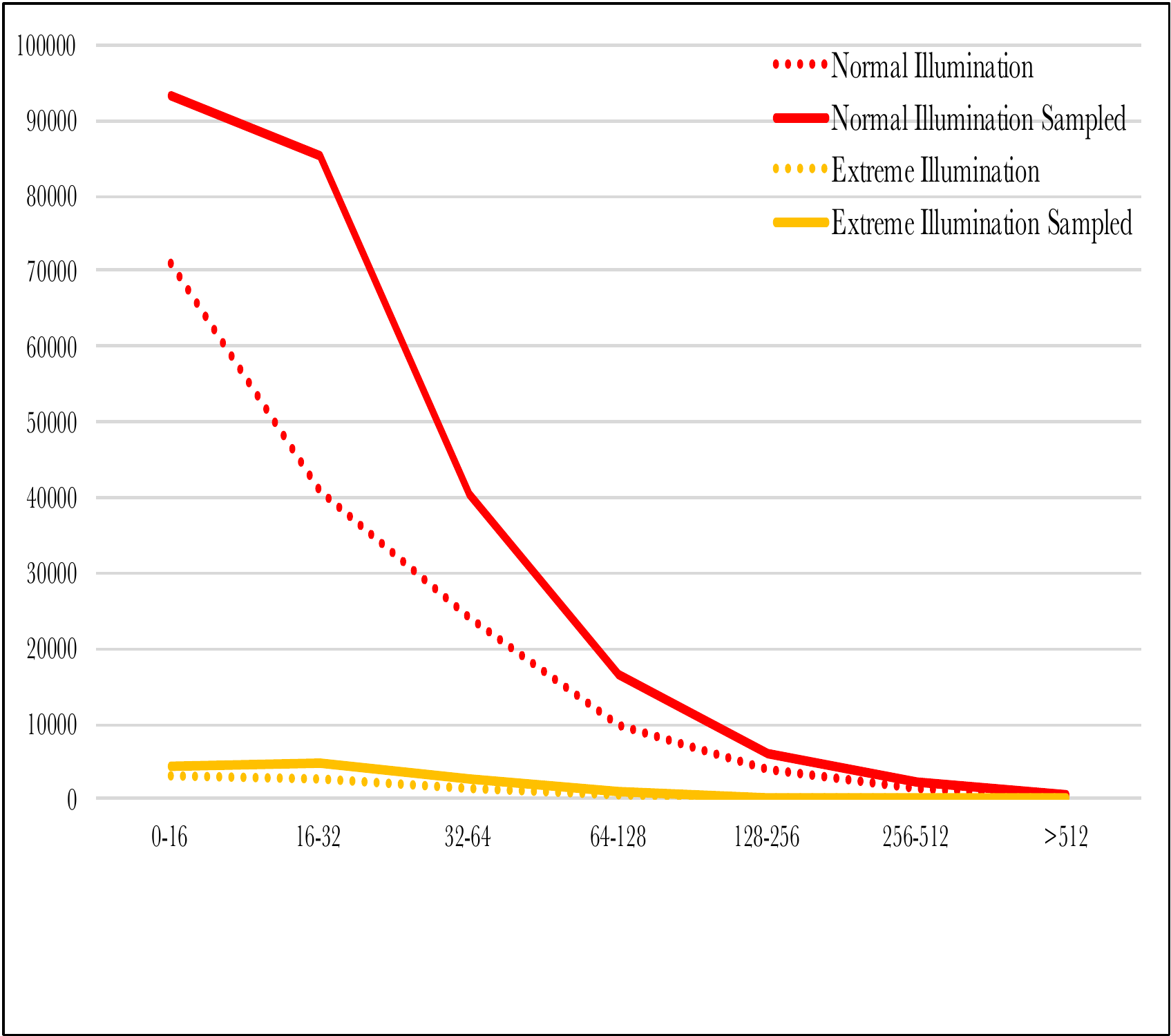}}\hfill
\caption{Data-anchor-sampling changes the distribution of the train data.
Dotted lines show the
distribution of certain attribute, while solid lines represent the corresponding distribution of
those attribute after the data-anchor-sampling.}
\label{fig:dataanchorsampling}
\end{figure}
\textbf{Data-anchor-sampling.}
Data sampling~\cite{Thompson2012}  is a classical subject in statistics, machine learning and pattern recognition,
it achieves great development in recent years. For the task of objection detection,
Focus Loss~\cite{Lin2017b} address the class imbalance by reshaping the standard cross entropy loss.

Here we utilize a  data augment sample method named Data-anchor-sampling.
In short, data-anchor-sampling resizes train images by reshaping a random face in this image to a random smaller anchor size.
More specifically, we first randomly select a face of size $s_{face}$ in a sample.
As previously mentioned that the scales of anchors in our PyramidBox,
as shown in Sec.~\ref{sec:LFPN},  are
\[
s_{i} = 2^{4+i}, \mbox{ for }i = 0, 1, \ldots, 5,
\]
let
\[
i_{anchor} = \mbox{argmin}_i \mbox{abs}(s_{anchor_i} - s_{face})
\]
be the index of the nearest anchor scale from the selected face, then we choose a random index $i_{target}$ in the set
\[
\{ 0, 1, \ldots, \mbox{min}(5, i_{anchor} + 1)\},
\]
finally, we resize the face of size of $s_{face}$ to the size of
\[
s_{target} = random(s_{i_{target}}/2,  s_{i_{target}}*2).
\]
Thus, we got the image resize scale
\[
s^* = s_{target}/s_{face}.
\]
By resizing the original image with the scale $s^*$
and cropping a standard size of $640\times640$ containing the selected face randomly, we get
the anchor-sampled train data.
For example, we first select a face randomly, suppose its size is $140$,
then its nearest anchor-size is $128$, then we need to choose a target size from
$16, 32, 64, 128$ and $256$.
In general, assume that we select $32$, then we resize the original
image by scale of $32/140 = 0.2285$.
Finally, by cropping a $640\times640$ sub-image from the last resized image containing
the originally selected face, we get the sampled train data.

As shown in Fig.~\ref{fig:dataanchorsampling},
data-anchor-sampling changes the distribution of the
train data as follows:
1) the proportion of small faces is larger than the large ones.
2) generate smaller face samples through larger ones to increase the diversity of face samples of smaller scales.

\textbf{PyramidBox loss.}
As a generalization of the multi-box loss in~\cite{Girshick2015},  we employ the \emph{PyramidBox Loss}
function for an image is defined as
\begin{equation}
\label{eq:loss_function}
L(\{p_{k,i}\}, \{t_{k,i}\})  = \sum_k \lambda_k L_k(\{p_{k,i}\}, \{t_{k,i}\}),
\end{equation}
where the $k$-th pyramid-anchor loss is given by
\begin{equation}
\label{eq:loss_part_function}
L_k(\{p_{k,i}\}, \{t_{k,i}\}) = \frac{\lambda}{N_{k,cls}}\sum_{i_k}L_{k,cls}(p_{k,i},p_{k,i}^*)
                              + \frac{1}{N_{k,reg}}\sum_{i_k}p_{k,i}^*L_{k,reg}(t_{k,i},t_{k,i}^*).
\end{equation}
Here $k$ is the index of pyramid-anchors  ($k = 0, 1$, and $2$ represents for
face, head and body, respectively, in our experiments), and $i$ is the index of an anchor and $p_{k,i}$ is the predicted
probability of anchor $i$ being the $k$-th object (face, head or body). The ground-truth label defined by
\begin{equation}
p_{k,i}^* = \left\{\begin{array}{lll}
            1, & \mbox{if the anchor down-sampled by stride ${s_{pa}}^k$ is positive},\\[7pt]
            0, & otherwise.
            \end{array}\right.
\end{equation}
For example, when $k=0$, the ground-truth label is equal to the label in Fast R-CNN~\cite{Girshick2015},
otherwise, when $k\ge1$, one can determine the corresponding label
by matching between the down-sampled anchors and ground-truth faces.
Moreover, $t_{k,i}$ is a vector representing
the $4$ parameterized coordinates of the predicted bounding box, and $t_{k,i}^*$ is that of ground-truth box
associated with a positive anchor,  we can define it by
\begin{equation}
\begin{array}{ll}
t_{k,i}^* = (&t_x^* + \frac{1-s_{p_a}^k}{2}t_w^*s_{w,k}+\Delta_{x,k}, t_y^* +\frac{1-s_{p_a}^k}{2}t_h^*s_{h,k}+\Delta_{y,k}, \\[7pt]
          &s_{p_a}^kt_w^*s_{w,k}-2\Delta_{x,k}, s_{p_a}^kt_h^*s_{h,k}-2\Delta_{y,k}),
\end{array}
\end{equation}
where ${\Delta_{x,k}}$ and ${\Delta_{y,k}}$ denote offset of shifts, 
$s_{w,k}$ and $s_{h,k}$ are scale factors respect to width and height 
respectively. 
In our experiments, we set $\Delta_{x,k}=\Delta_{y,k}=0,s_{w,k}=s_{h,k}=1$ for $k<2$ 
and ${\Delta_{x,2}}=0,{\Delta_{y,2}}=t_h^*,s_{w,2}=\frac{7}{8},s_{h,2}=1$ for $k =2$.
The classification loss $L_{k,cls}$ is log loss over two classes ( face \emph{vs.} not face) and
the regression loss $L_{k,reg}$ is the smooth $L_1$ loss defined in~\cite{Girshick2015}.
The term $p_{k,i}^*L_{k,reg}$ means the regression loss is activated only for positive anchors and disabled
otherwise. The two terms are normalized with $N_{k,cls}$, $N_{k,reg}$, and balancing weights $\lambda$ and
$\lambda_k$ for $k = 0, 1, 2$.

\textbf{Optimization.} As for the parameter initialization, our PyramidBox use the pre-trained parameters
from VGG$16$~\cite{Russakovsky2015}.
The parameters of \emph{conv$\_$fc\,$6ß¬7$} and \emph{conv$\_$fc\,$7$} are initialized
by sub-sampling parameters from \emph{fc\,$6$} and \emph{fc\,$7$}
of VGG$16$ and the other additional layers are randomly initialized with ``xavier" in~\cite{Glorot2010}.
We use a learning rate of $10^{-3}$ for $80$k iterations, and $10^{-4}$ for the next $20$k iterations,
and $10^{-5}$ for the last $20$k iterations on the WIDER FACE training set with batch size 16.
We also use a momentum of $0.9$ and a weight decay of $0.0005$~\cite{Krizhevsky2012}.

\section{Experiments}
\label{sec:exper}
In this section, we firstly analyze the effectiveness of our PyramidBox through a set of experiments,
and then evaluate the final model on WIDER FACE and FDDB
face detection benchmarks.

\subsection{Model Analysis}
We analyze our model on the WIDER FACE validation set by contrast experiments.

\textbf{Baseline.} Our PyramidBox shares the same architecture of S$^3$FD, so we directly use it as a baseline.

\textbf{Contrast Study.}
To better understand PyramidBox, we conduct contrast experiments to evaluate the contributions
of each proposed component, from which we can get the following conclusions.

\textbf{Low-level feature pyramid network (LFPN) is crucial for detecting hard faces.}
The results listed in Table~\ref{table:lfpn} prove that LFPN started from a middle layer,
using \emph{conv\,$\_$fc$7$} in our PyramidBox,
is more powerful, which implies that features with large gap in scale may not help each other.
The comparison between the first and forth column of Table~\ref{table:lfpn} indicates that
LFPN increases the mAP by $1.9\%$ on hard subset.
This significant improvement demonstrates the effectiveness of joining high-level semantic features with the low-level ones.

\begin{table}[h]
\begin{center}
\caption{Performances of LFPN starting from different layers.}
\label{table:lfpn}
\begin{tabular}{ll||c|c|c|c|c|c}
\hline
\multicolumn{2}{c||}{Start layer} & Baseline &\emph{conv\,$7\_2$}(\textbf{FPN}) & \emph{conv\,$6\_2$} & \emph{conv\,$\_$fc$7$}(\textbf{LFPN}) & \emph{conv\,$5\_3$} & \emph{conv\,$4\_3$}\\
\hline
\hline
\multicolumn{2}{c||}{RF/InputSize} & & $1.13125$ & $0.73125$ & $0.53125$ & $0.35625$ & $0.16875$\\
\hline
\multirow{3}*{mAP} & easy   & $94.0$ & $93.9$ & $94.1$ & $\bm{94.3}$ & $94.1$ & $93.6$\\
                 ~ & medium & $92.7$ & $92.9$ & $93.1$ & $\bm{93.3}$ & $93.1$ & $92.5$\\
                 ~ & hard   & $84.2$ & $85.9$ & $85.9$ & $\bm{86.1}$ & $85.7$ & $84.8$ \\
\hline
\end{tabular}
\end{center}
\end{table}

\textbf{Data-anchor-sampling makes detector easier to train.} We employ Data-anchor-sampling based on LFPN network and the result shows that our data-anchor-sampling
effectively improves the performance. The mAP is increased by $0.4\%$, $0.4\%$ and $0.6\%$ on easy, medium
and hard subset, respectively. One can see that Data-anchor-sampling
works well not only for small hard faces, but also for easy and medium faces.

\textbf{PyramidAnchor and PyramidBox loss is promising.}
\begin{table}[h]
\begin{center}
\caption{The Parameters of PyramidAnchors.}
\label{table:pyramidanchors_param}
\begin{tabular}{ll||c|c|c|c|c}
\hline
\multicolumn{2}{c||}{\textbf{Method}} & Baseline & $(K, s_{pa})$ & $(K, s_{pa})$ & $(K, s_{pa})$ & $(K, s_{pa})$\\
\multicolumn{2}{c||}{} & & $(1, 1.5)$ & $(1, 2.0)$ & $(1, 3.0)$ & $(2, 2.0)$\\
\hline
\hline
\multirow{3}*{mAP} & easy   & $94.0$ & $93.8$ & $94.2$ & $94.3$ & $\bm{94.7}$\\
                 ~ & medium & $92.7$ & $92.7$ & $93.0$ & $93.1$ & $\bm{93.3}$\\
                 ~ & hard   & $84.2$ & $84.8$ & $84.9$ & $85.0$ & $\bm{85.1}$\\
\hline
\end{tabular}
\end{center}
\end{table}
By comparing the first and last column in Table~\ref{table:pyramidanchors_param},
one can see that PyamidAnchor effectively improves the
performance, i.e., $0.7\%$, $0.6\%$ and $0.9\%$ on easy, medium and hard, respectively.
This dramatical improvement shows that learning contextual information is helpful to the task of detection,
especially for hard faces.

\textbf{Wider and deeper context prediction module is better.}
Table~\ref{table:cpm_res} shows that the performance of CPM is better than both DSSD module and SSH context module. Notice that the combination of SSH and DSSD gains very little compared to SSH alone, which indicates that large receptive field is more important to predict the accurate location and classification.
In addition, by comparing the last two column of Table~\ref{table:ablative}, one can find that
the method of Max-in-out improves the mAP on WIDER FACE validation set
about $+0.2\%$(Easy), $+0.3\%$(Medium) and $+0.1\%$(Hard), respectively.
\begin{table}[h]
\begin{center}
\caption{Context-sensitive Predict Module.}
\label{table:cpm_res}
\begin{tabular}{ll||c|c|c}
\hline
\multicolumn{2}{c||}{\textbf{Method}} & DSSD prediction module & SSH context module & \textbf{CPM} \\
\hline
\hline
\multirow{3}*{mAP} & easy   & $95.3$ & $95.5$ & $\bm{95.6}$ \\
                 ~ & medium & $94.3$ & $94.3$ & $\bm{94.5}$ \\
                 ~ & hard   & $88.2$ & $88.4$ & $\bm{88.5}$ \\
\hline
\end{tabular}
\end{center}
\end{table}

To conclude this section, we summarize our results in Table~\ref{table:ablative},
from which one can see that mAP increase $2.1\%$, $2.3\%$ and $\bm{4.7\%}$ on easy, medium and \textbf{hard} subset, respectively.
This sharp increase demonstrates the effectiveness of proposed PyramidBox, especially for hard faces.
\begin{table}[h]
\begin{center}
\caption{Contrast results of the PyramidBox on WIDER FACE validation subset.}
\label{table:ablative}
\begin{tabular}{rc||c|cccc|c}
\hline
\multicolumn{2}{c||}{Contribution}                         & Baseline &   &  & & &\textbf{PyramidBox}\\
\hline
\hline
\multicolumn{2}{r||}{lfpn?}                    &  & ~$\surd$  & ~$\surd$ & ~$\surd$ & ~$\surd$ & ~$\surd$\\
\hline
\multicolumn{2}{r||}{data-anchor-sampling?}    &  &  & ~$\surd$ & ~$\surd$ & ~$\surd$ & ~$\surd$\\
\hline
\multicolumn{2}{r||}{pyramid-anchors?}         &  &   &   &  ~$\surd$ & ~$\surd$ & ~$\surd$\\
\hline
\multicolumn{2}{r||}{context-prediect-module?} &  &   &   & & ~$\surd$ & ~$\surd$\\
\hline
\multicolumn{2}{r||}{max-in-out?}              &  &   &   & &  & ~$\surd$\\
\hline
    & easy   & $94.0$ & ~$94.3$ & ~$94.7$ & ~$95.5$ & ~$95.9$ & ~$\bm{96.1}$ \\
mAP & medium & $92.7$ & ~$93.3$ & ~$93.7$ & ~$94.3$ &~$94.7$ & ~$\bm{95.0}$\\
    & hard   & $84.2$ & ~$86.1$ & ~$86.7$ & ~$88.3$ &~$88.8$ & ~$\bm{88.9}$\\
\hline
\end{tabular}
\end{center}
\end{table}

\subsection{Evaluation on Benchmark}
We evaluate our PyramidBox on the most popular face detection benchmarks, including
Face Detection Data Set and Benchmark (FDDB)~\cite{Jain2010} and
WIDER FACE~\cite{Yang2015b}.

\textbf{FDDB Dataset.}
\begin{figure}[t]
\centering
\subfloat[Discontinous ROC curves] {%
\includegraphics[height=3.8cm]{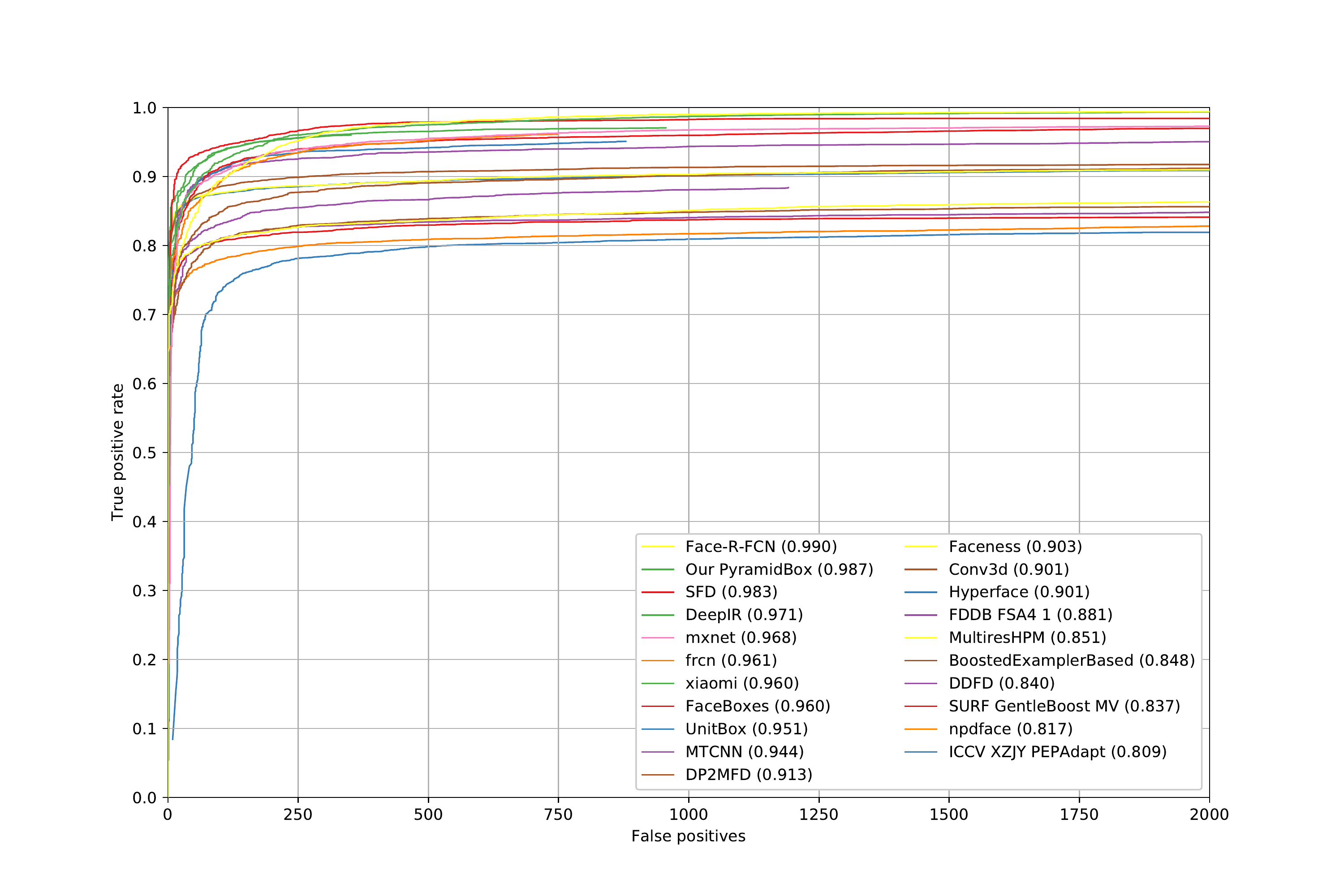}}\hfill
\subfloat[Continous ROC curves] {%
\includegraphics[height=3.8cm]{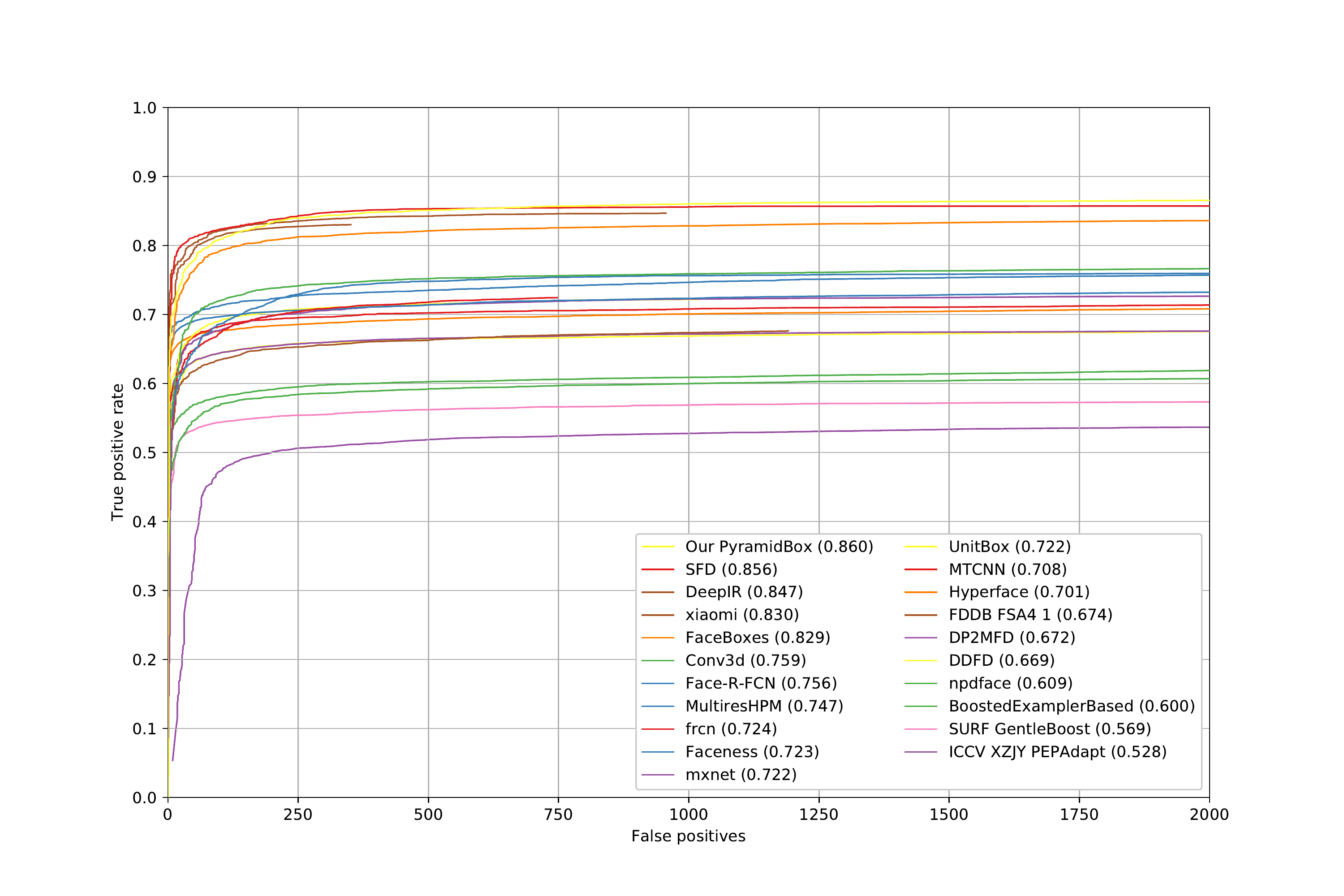}}\hfill
\caption{Evaluation on the FDDB dataset.}
\label{fig:fddbres}
\end{figure}
It has $5,171$ faces in $2,845$ images collected from the Yahoo! news website.
We evaluate our face detector on FDDB against the
other state-of-art methods~\cite{Zhang2017,Liao2016,Zhang2016,Zhang2017b,Yu2016,Barbu14,Triantafyllidou2016,Yang2015,Li2016,Farfade2015,Ghiasi2015,Kumar2015,Li2013,Li2013b,Li2014,Ohnbar2016,Ranjan2015,Ranjan2016,Cai2016,Wan2016}. The PyramidBox achieves state-of-art performance and the result is shown in Fig.~\ref{fig:fddbres}(a) and~Fig.~\ref{fig:fddbres}(b).

\textbf{WIDER FACE Dataset.}
\begin{figure}[t]
\centering
\subfloat[Val: Easy] {%
\includegraphics[height=3.0cm]{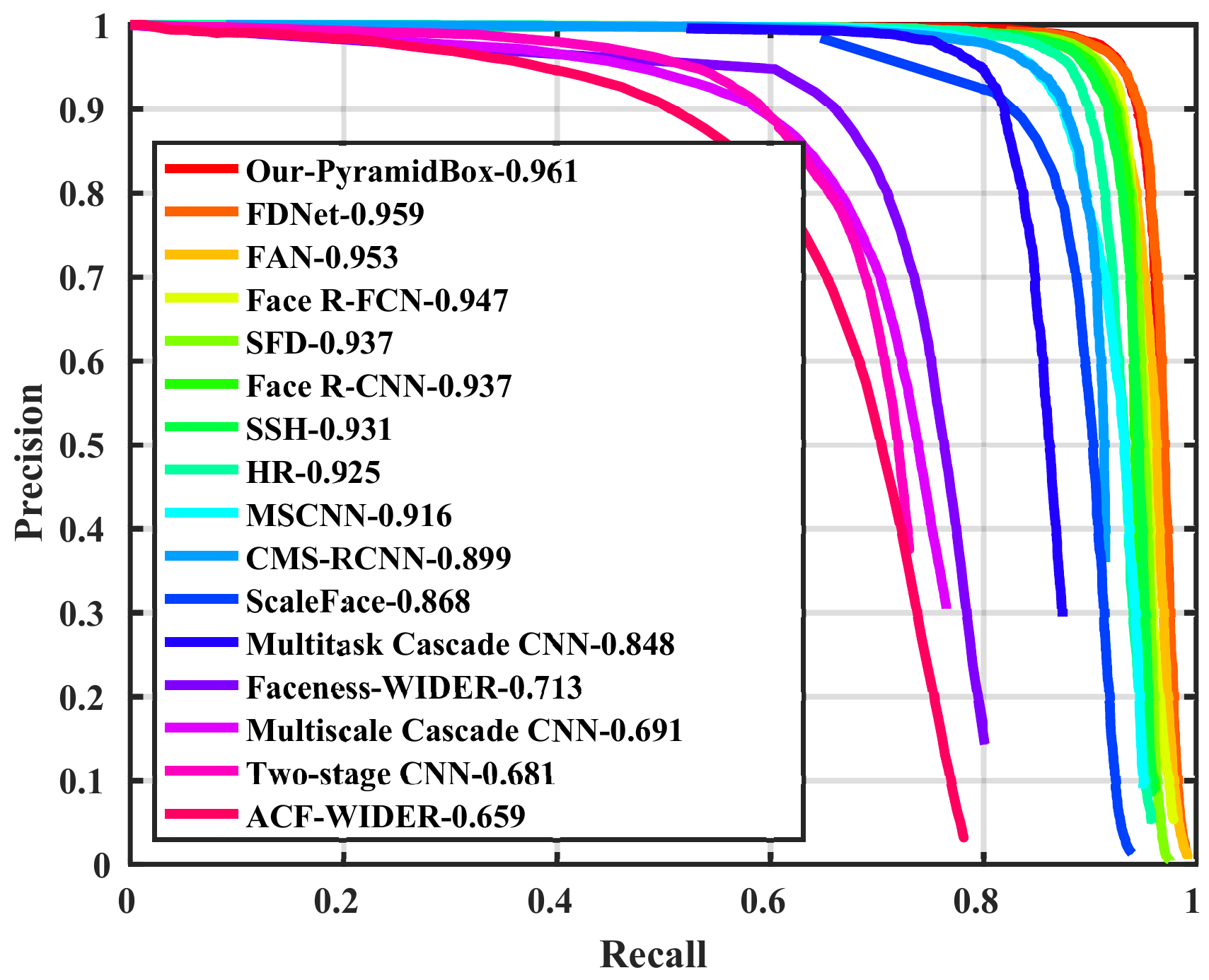}}\hfill
\subfloat[Val: Medium] {%
\includegraphics[height=3.0cm]{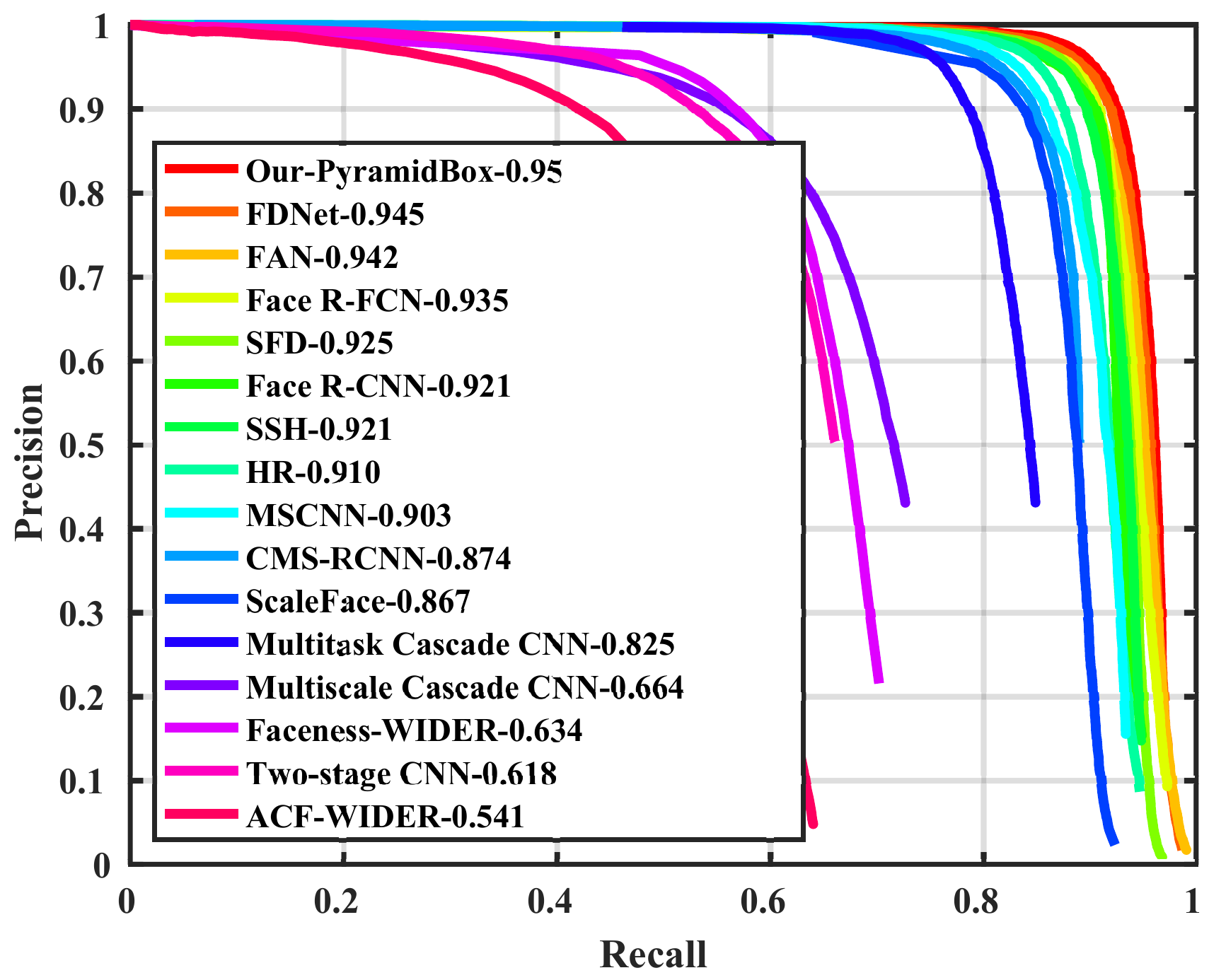}}\hfill
\subfloat[Val: Hard] {%
\includegraphics[height=3.0cm]{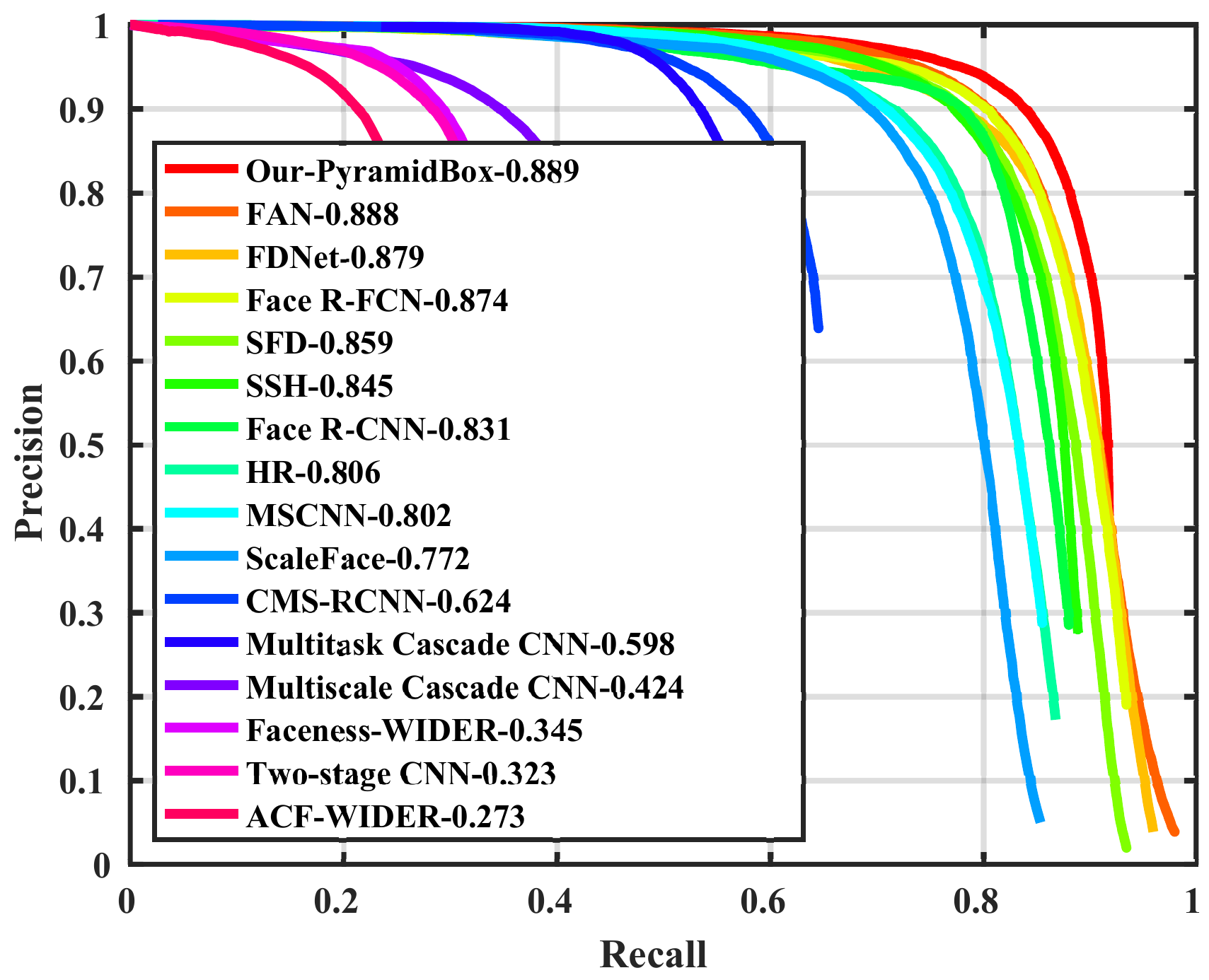}}\hfill
\subfloat[Test: Easy] {%
\includegraphics[height=3.0cm]{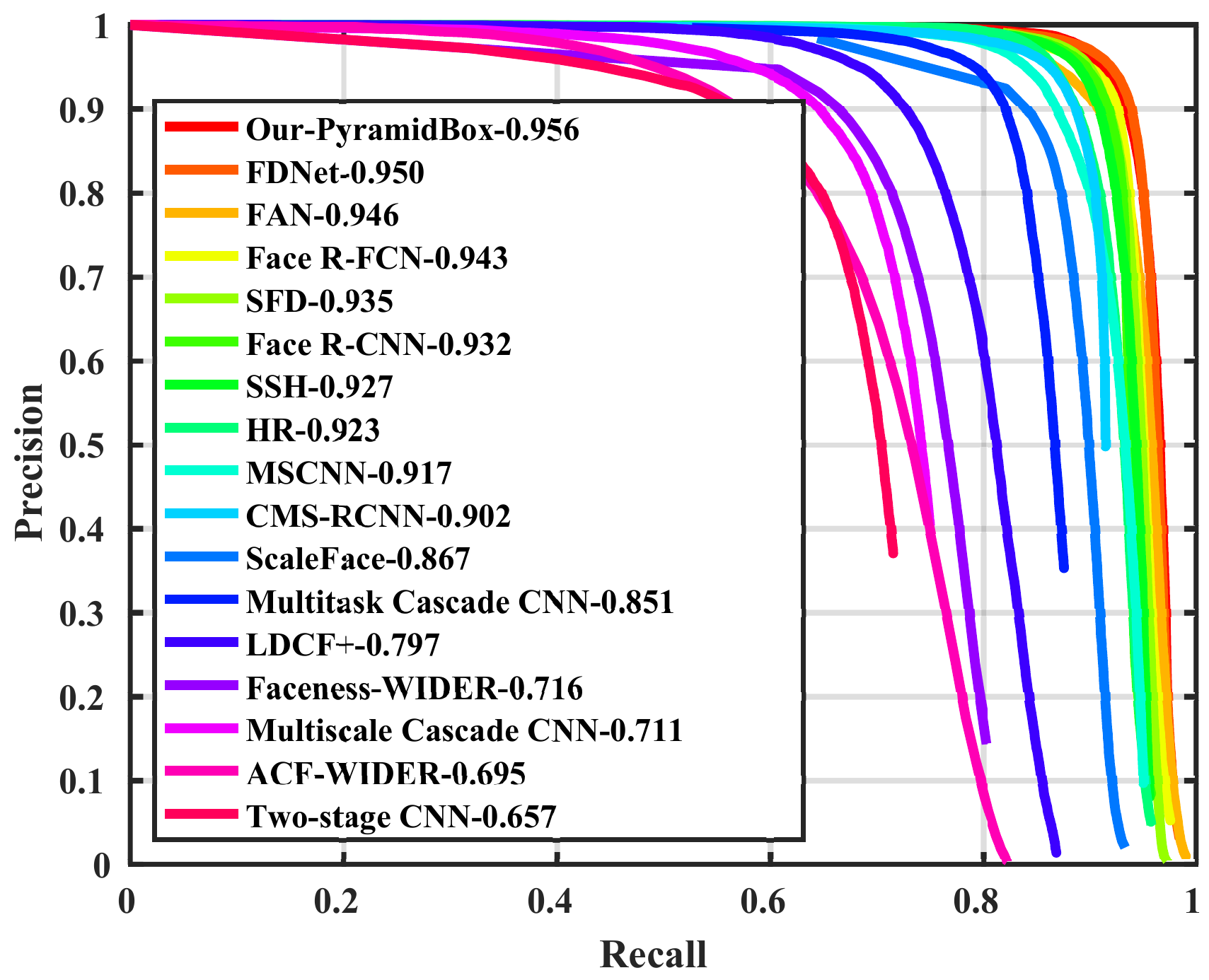}}\hfill
\subfloat[Test: Medium] {%
\includegraphics[height=3.0cm]{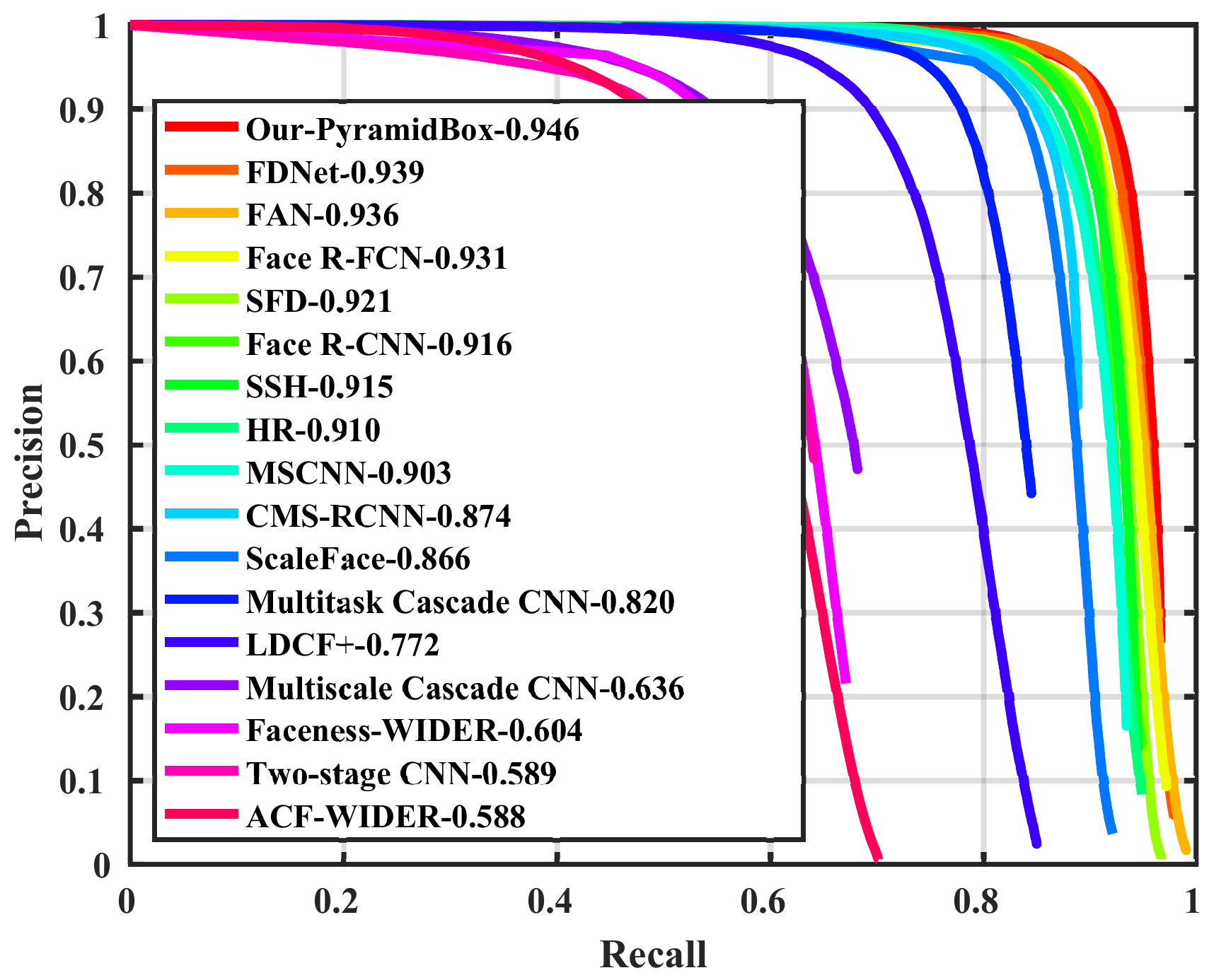}}\hfill
\subfloat[Test: Hard] {%
\includegraphics[height=3.0cm]{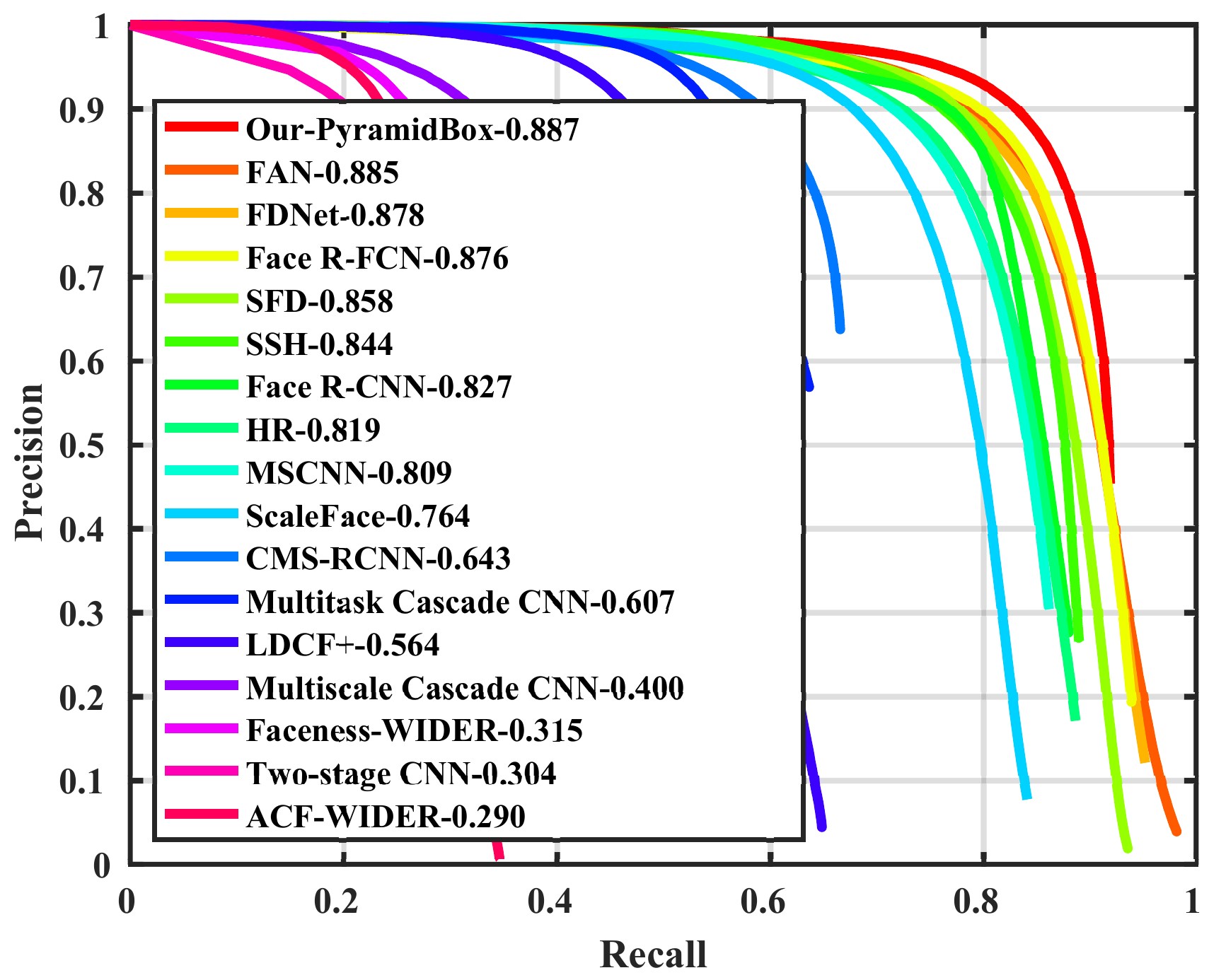}}\hfill
\caption{Precision-recall curves on WIDER FACE validation and test sets.}
\label{fig:widerface}
\end{figure}
It contains $32,203$ images and $393,703$ annotated faces with a high degree of variability in scale, pose and occlusion.
The database is split into training ($40\%$), validation ($10\%$) and testing ($50\%$) set,
where both validation and test set are divided into ``easy", ``medium" and``hard" subsets,
regarding the difficulties of the detection.
Our PyramidBox is trained only on the training set and evaluated on both validation set
and testing set comparing with the state-of-the-art face detectors,
such as~\cite{Zhang2017,Yang2014,Zhang2016,Zhang2017b,Yang2015,Zhu2016,Yang2015b,Ohnbar2016,Zhang2018,Wang2017b,Cai2016,Wang2017,Najibi2017,Yang2017,Hu2017,wang2017c}.
Fig.~\ref{fig:widerface} presents the precision-recall curves and mAP values.
Our PyramidBox outperforms others across all three subsets,
i.e. $0.961$ (easy), $0.950$ (medium), $0.889$ (hard)
for validation set, and $0.956$ (easy), $0.946$ (medium), $0.887$ (hard) for testing set.

\section{Conclusion}
\label{sec:conclu}
This paper proposed a novel context-assisted single shot face detector,
denoted as PyramidBox, to handle the unconstrained face detection problem.
We designed a novel context anchor, named PyramidAnchor, to supervise face detector to
learn features from contextual parts around faces. Besides, we modified feature pyramid
network into a low-level feature pyramid network to combine features from high-level and
high-resolution, which are effective for finding small faces. We also proposed a wider
and deeper prediction module to make full use of joint feature. In addition, we introduced Data-anchor-sampling
to augment the train data to increase the diversity of train data for small faces.
The experiments demonstrate
that our contributions lead PyramidBox to the state-of-the-art performance on the common
face detection benchmarks, especially for hard faces.

\vspace{.2cm} \noindent{\bf Acknowledgments.}
 We wish to thank Dr. Shifeng Zhang and Dr. Yuguang Liu for many helpful discussions.
\clearpage


\clearpage

\section*{Appendix}

\begin{figure}
\centering
\includegraphics[height=6.3cm]{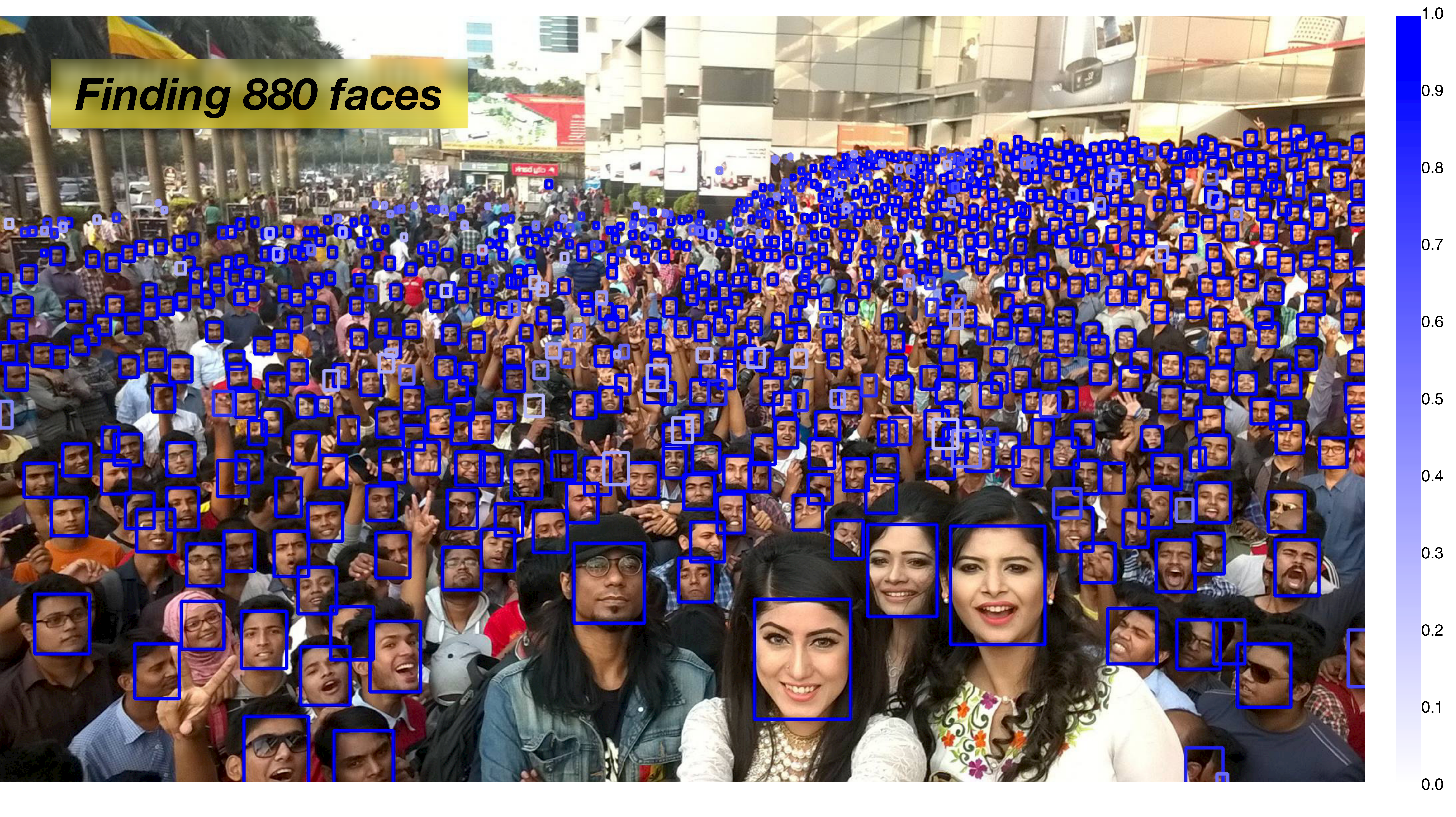}
\caption{An exampler which has extreme variability in the face regions. Our PyramidBox can find 880 faces out of the reportedly 1000 present in the above image. On the right of image, detector confidence is present to you directly by colorbar. Please zoom in for more details.}
\label{fig:demo_img}
\end{figure}

\begin{figure}[t]
\centering
\includegraphics[height=18cm]{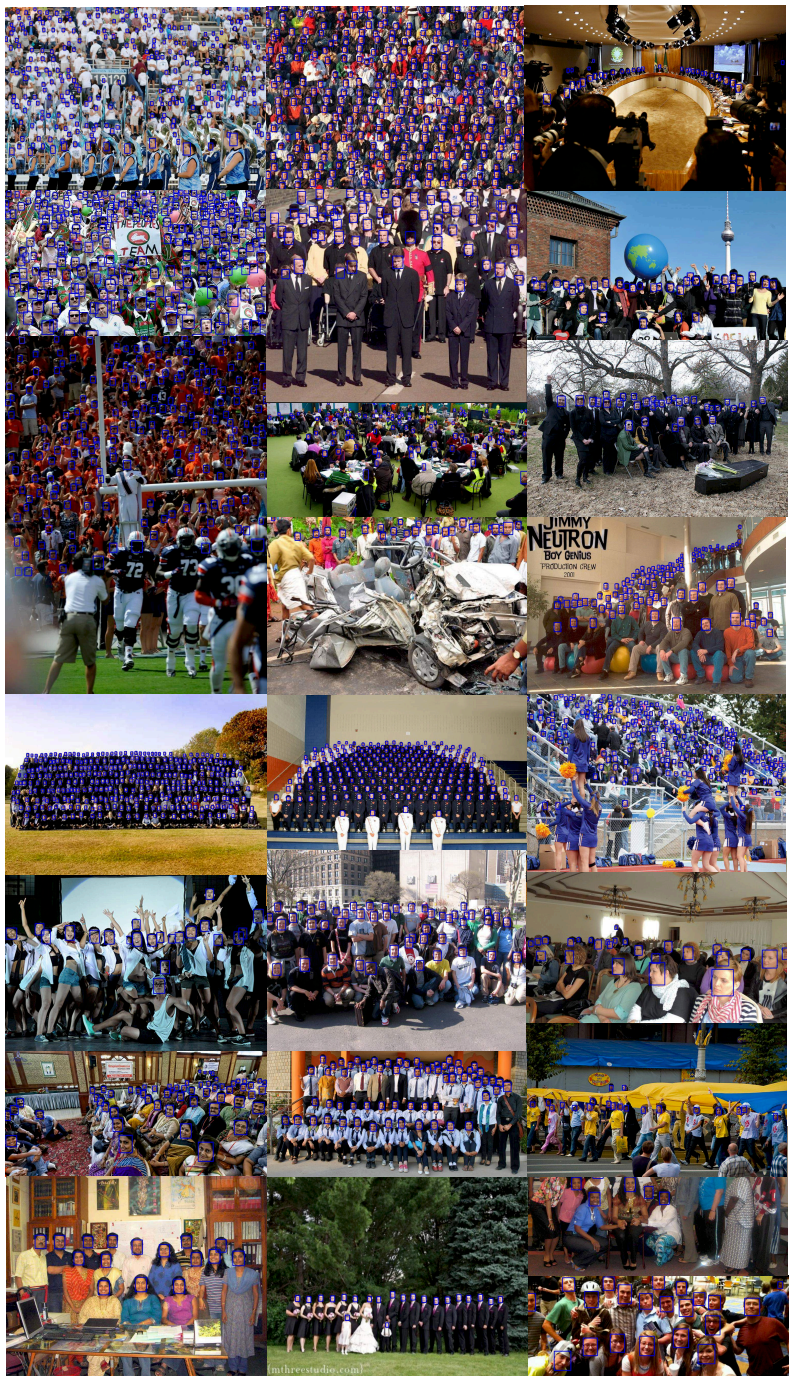}
\caption{Our PyramidBox can handle faces with a wide range of face scales. Blue represent the detector confidence above 0.8.}
\label{fig:robustness_scale}
\end{figure}

\begin{figure}
\centering
\includegraphics[height=9cm]{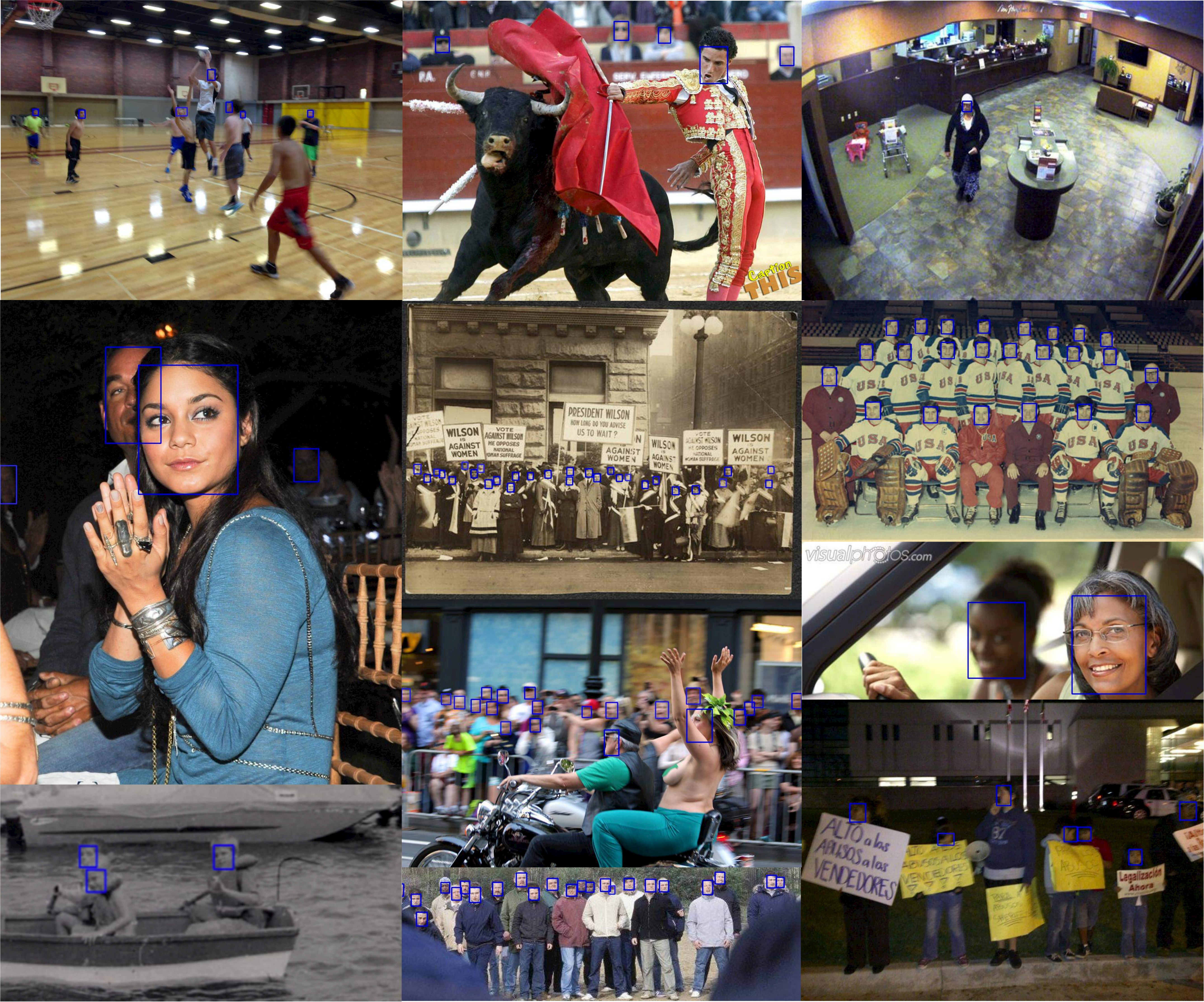}
\caption{Our PyramidBox is able to handle various forms of blur, a key factor leading to the degradation of image quality. Blue represent the detector confidence above 0.8.}
\label{fig:robustness_blur}
\end{figure}

\begin{figure}
\centering
\includegraphics[height=6.5cm]{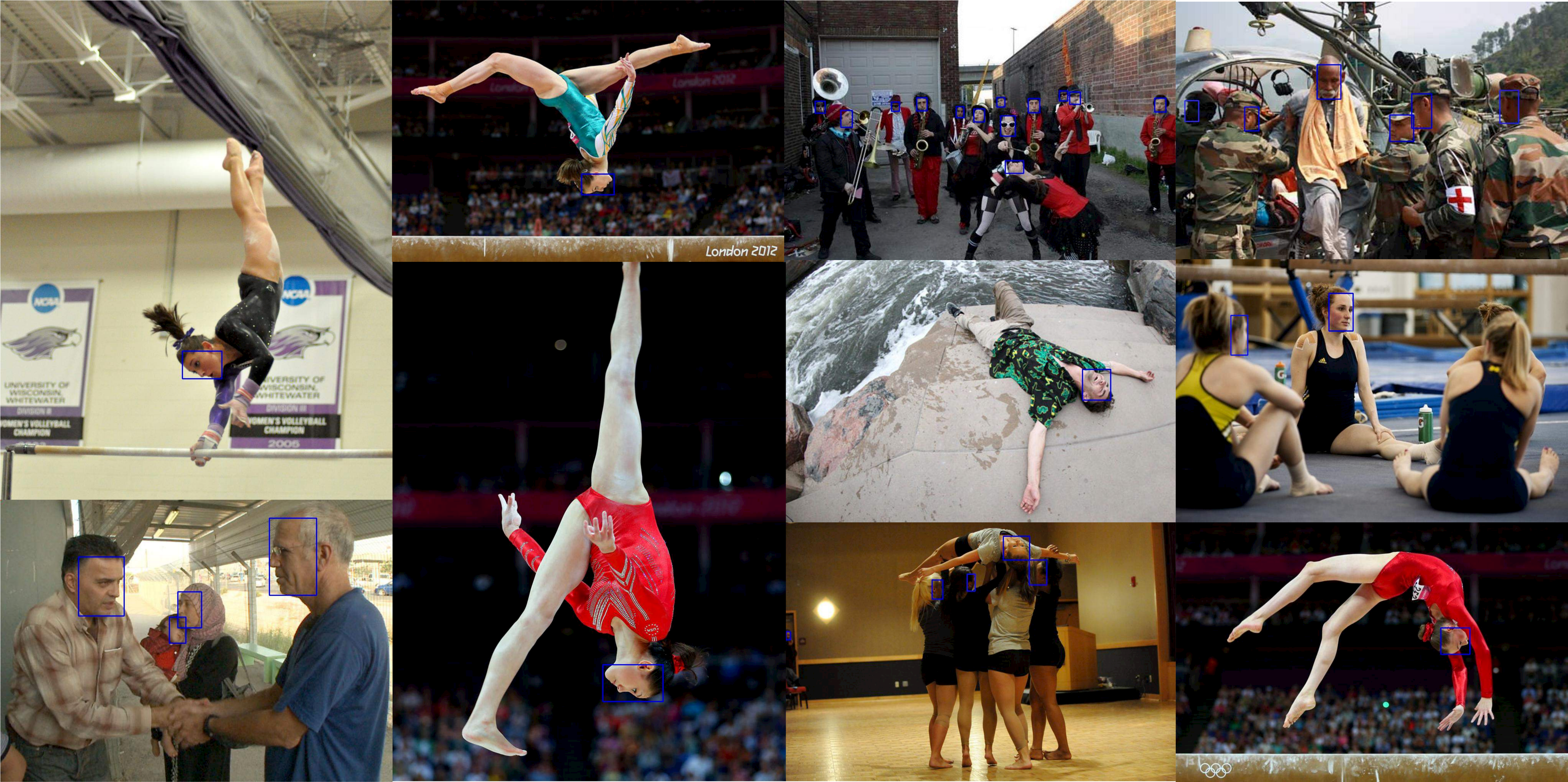}
\caption{The results of our PyramidBox across pose is shown in this figure, and blue represent the detector confidence above 0.8.}
\label{fig:robustness_pose}
\end{figure}

\begin{figure}[t]
\centering
\includegraphics[height=12cm,width=13cm]{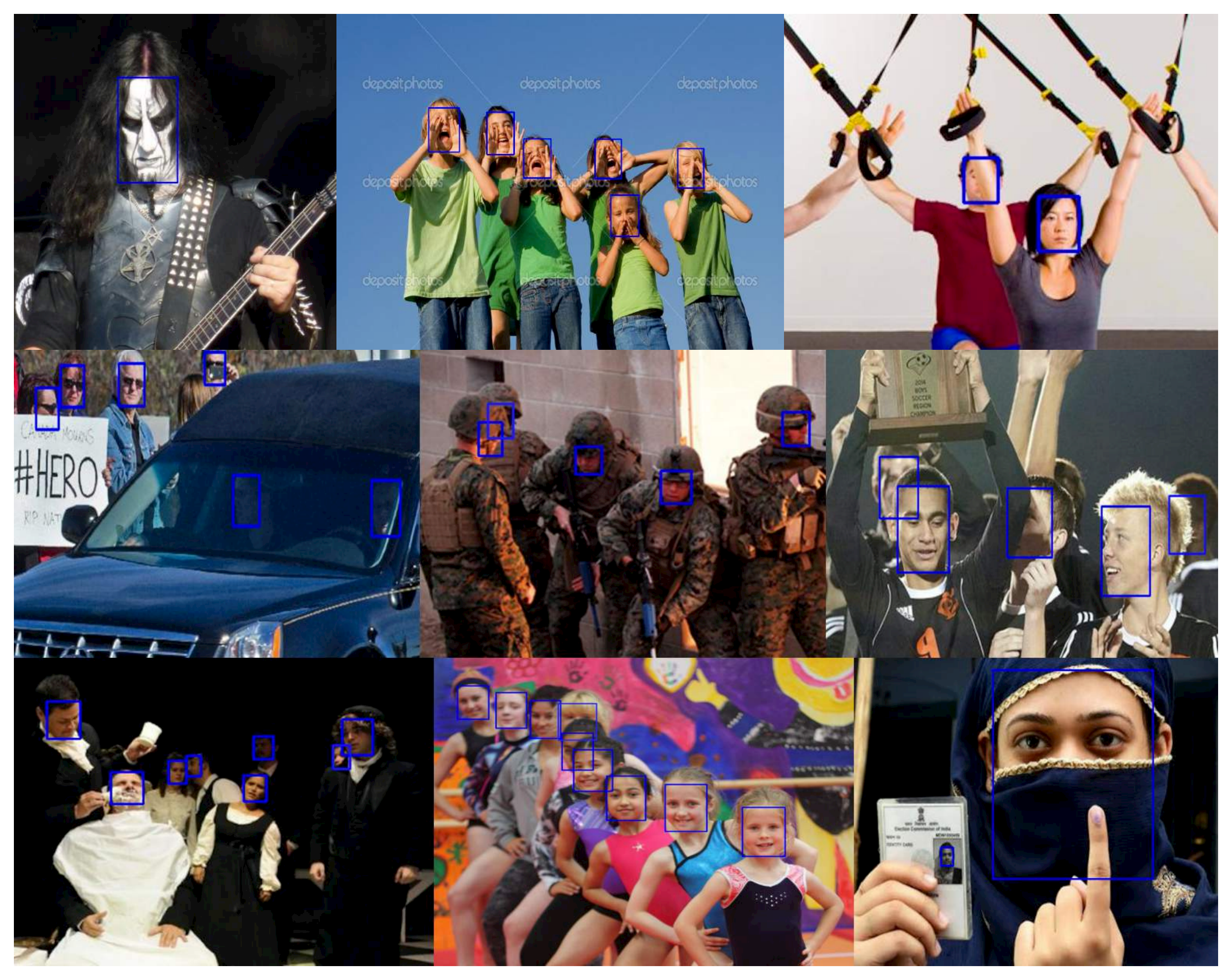}
\caption{Our PyramidBox can handle facial occlusions caused by sunglasses, mask, hat etc., and blue represent the detector confidence above 0.8.}
\label{fig:robustness_occlusion}
\end{figure}

In this section, we show the robustness of our PyramidBox algorithm by testing it on the face images having large variance in scale, blur, pose and occlusion. Even in the images filled with small, blurred or partially occluded faces and big face with exaggerate expression, our PyramidBox can recall most of these faces, see Fig.~\ref{fig:demo_img}.  Besides, the robustness of scale, occlusion, blur, and pose is respectively described in the Fig.~\ref{fig:robustness_scale}, Fig.~\ref{fig:robustness_blur}, Fig.~\ref{fig:robustness_pose} and Fig.~\ref{fig:robustness_occlusion}.

\end{document}